\documentclass[letterpaper]{article} 
\usepackage{aaai2027}  
\usepackage[hyphens]{url}  
\usepackage{graphicx} 
\usepackage{natbib}  
\usepackage{caption} 
\usepackage{algorithm}
\usepackage{algorithmic}
\usepackage{amssymb}
\usepackage{amsmath} 
\usepackage{subcaption}
\usepackage{newfloat}
\usepackage{listings}
\DeclareCaptionStyle{ruled}{labelfont=normalfont,labelsep=colon,strut=off} 
\floatstyle{ruled}
\newfloat{listing}{tb}{lst}{}
\floatname{listing}{Listing}

\usepackage{booktabs}

\newcommand{\best}[1]{\textbf{\color{red}#1}}
\newcommand{\second}[1]{\textit{\color{blue}#1}}

\title{Learning Ordinal Degradation Representations with Textual Priors for Diffusion-Based Blind Image Super-Resolution}
\author{
    Yi-Cheng Liao\textsuperscript{\rm 1},\qquad
    Shyang-En Weng\textsuperscript{\rm 1},\qquad
    Yu-Syuan Xu\textsuperscript{\rm 2},\qquad
    Chia-Hung Yuan\textsuperscript{\rm 2},\\
    Wei-Chen Chiu\textsuperscript{\rm 1},\qquad
    Ching-Chun Huang\textsuperscript{\rm 1}\thanks{Corresponding author (chingchun@nycu.edu.tw)}\\
}
\affiliations{
    \textsuperscript{\rm 1}National Yang Ming Chiao Tung University, Taiwan,\qquad
    \textsuperscript{\rm 2}MediaTek Inc., Taiwan\\

}

\begin{document}

\maketitle

\begin{abstract}
Blind image super-resolution (Blind SR) has achieved remarkable perceptual quality via generative priors. 
However, lacking clear degradation representations such as varying severity and mixtures, these methods fail to accurately reflect the actual degradation process. 
This limitation severely compromises restoration fidelity and leads to content inconsistencies, especially in diffusion-based blind SR models that rely on simple textual descriptions for contextual guidance.
To bridge the gap between high-level semantics and low-level degradation artifacts, we introduce Ordinal Degradation CLIP (OD-CLIP), leveraging textual priors to enhance the learning of continuous degradation-level representations. 
Unlike standard CLIP text encoders, which struggle to represent numerical intensity, OD-CLIP moves beyond coarse labels by modeling unknown degradations as a continuous spectrum representing quality. 
By learning an ordinal embedding from low-quality inputs, our design captures both degradation types and their relative severity, explicitly modeling the degradation hierarchy and enabling interpolation across unseen levels. 
In our experiments, the OD-CLIP representation demonstrates stronger ordinal ranking and perceptual distance modeling compared to baseline methods. 
When applied to blind SR, we show that conditioning on OD-CLIP maintains fidelity and preserves content structures over existing methods in both unknown and mixed-degradation settings on real-world benchmarks.
\end{abstract}


\section{Introduction}
\label{sec:intro}
Image super-resolution (SR) aims to reconstruct high-resolution (HR) images with faithful textures and structures from low-resolution (LR) observations. Conventional single-image super-resolution (SISR) methods~\cite{VDSR,EDSR,SRGAN} typically rely on fixed, simplified degradation assumptions-most commonly ideal bicubic downsampling. Consequently, they suffer severe performance drops when applied to real-world images affected by complex, unknown factors such as blur, noise, and compression artifacts. To bridge this gap, blind super-resolution (Blind SR) seeks to restore real-world degraded inputs without requiring prior knowledge of the underlying degradation process. Early blind SR approaches~\cite{KernelGAN,DASR_cvpr,DASR_eccv,IKC} focused on explicitly estimating degradation representations (e.g., blur kernels or degradation maps) to guide the restoration process. This explicit modeling improves robustness to known degradation types, but the resulting representations remain too coarse to guide the synthesis of fine, highly realistic details under severe degradations.

\begin{figure}[!t]
    \centering
    \includegraphics[width=1.0\linewidth]{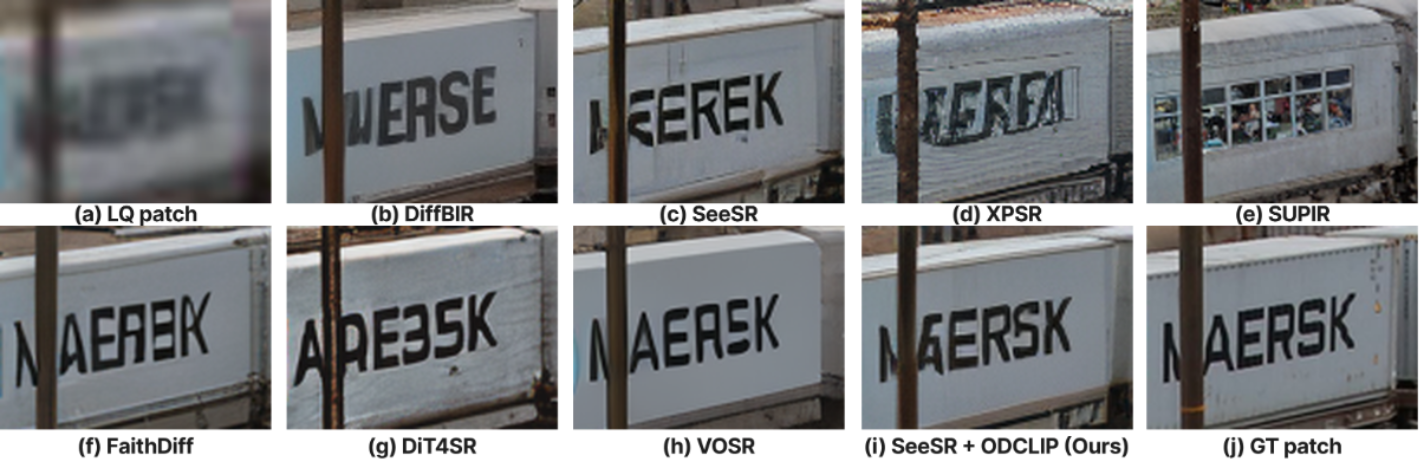}
    \caption{
    The SOTA SR methods in (b)–(h) fail to maintain content consistency and produce incorrect words and distorted structures. With OD-CLIP, our method in (i) faithfully recovers “MAERSK.”
    }
    \label{fig:intro_visualize}
\end{figure}

Driven by powerful generative priors, recent blind SR methods have achieved remarkable perceptual quality. Several state-of-the-art approaches~\cite{stablesr,pasd,diffbir,seesr,faithdiff} integrate vision-language models to provide rich semantic textual prompts alongside the low-resolution (LR) input. These methods excel at synthesizing realistic textures, yet this very strength can work against structural accuracy: when the conditioning signals are insufficient to strictly constrain the reverse diffusion process, the models tend to generate content that is locally plausible yet semantically inconsistent with the underlying scene structure—a failure mode we term \emph{content inconsistency}, as shown in Fig.~\ref{fig:intro_visualize}.

\begin{figure*}[!t]
    \centering
    \includegraphics[width=0.9\linewidth]{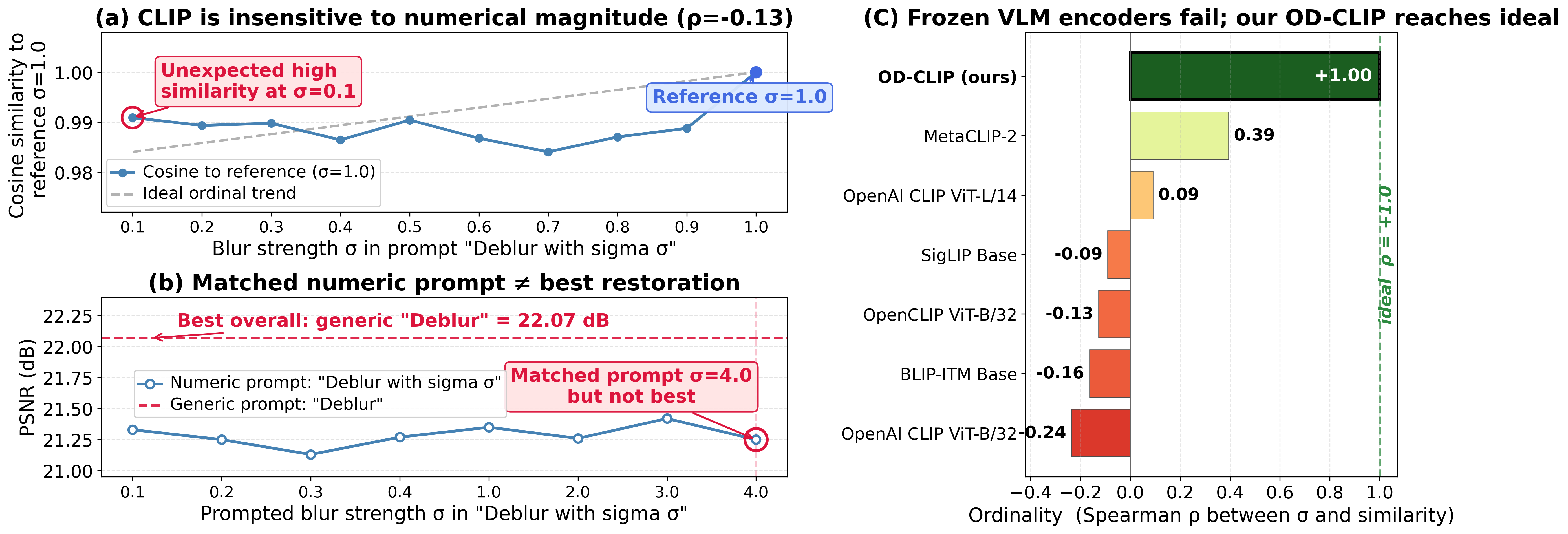}
    \caption{\textbf{Numerical insensitivity in frozen VLM encoders leads to unreliable restoration guidance.} (a) \textbf{Numerical insensitivity in CLIP.} Cosine similarity between CLIP prompt embeddings for different blur strengths $\sigma$ and the reference prompt at $\sigma=1.0$. The non-monotonic relationship ($\rho=-0.13$), such as the unexpectedly higher similarity at $\sigma=0.1$ than at $\sigma=0.9$, shows that CLIP does not reliably preserve the ordinal relationship between numerical values. (b) \textbf{Impact on image restoration.} PSNR of FaithDiff~\cite{faithdiff} on a fixed blurred input with $\sigma=4.0$, conditioned on numeric prompts specifying different blur strengths. The matched prompt ($\sigma=4.0$) does not yield the best restoration result and is outperformed by the generic prompt (``Deblur''), demonstrating that numerical cues from the frozen text encoder provide unreliable guidance for diffusion-based restoration. (c) \textbf{Ordinality across text encoders.} Spearman correlation between the prompted blur strength and its embedding similarity to the reference prompt. Existing frozen vision-language text encoders exhibit only weak or negative ordinality, whereas our OD-CLIP achieves the ideal correlation of $\rho=1.0$.}
    \label{fig:clip_problem}

\end{figure*}

While content inconsistency is frequently attributed to ambiguous semantic prompts, fundamental principles in blind SR dictate that restoration fidelity is intrinsically tied to the accuracy of degradation modeling. Inaccurate degradation estimation is closely linked to structural distortions and artifacts, suggesting that the degradation representation itself serves as an indispensable constraint for faithful reconstruction. Motivated by this, recent generative works have sought to integrate degradation-aware guidance to regulate the denoising process. 
For instance, DA-CLIP~\cite{daclip, wild_clip} encodes degradation types as categorical embeddings via cross-attention, whereas SPIRE~\cite{spire} uses numerical prompts (e.g., ``Deblur with sigma 4.0'') to specify degradation severity. However, neither reliably represents both type and severity: categorical embeddings are discrete, while frozen vision-language encoders are numerically insensitive and fail to preserve numerical magnitude or order. Consequently, increasing degradation levels can yield non-monotonic embedding similarities across multiple encoders~\cite{metaclip2, siglip, blip, openclip} (Fig.~\ref{fig:clip_problem}(a,c)). This ambiguity also undermines restoration conditioning: for a fixed degraded input, the matched numerical prompt performs no better than mismatched or generic prompts (Fig.~\ref{fig:clip_problem}(b)).

In this work, we revisit degradation modeling in blind SR from a vision-language alignment perspective. Rather than conditioning restoration on coarse textual prompts that cannot express degradation severity, we disentangle degradation into discrete types and continuous severity, and learn an ordinal representation over both. Our main contributions are summarized as follows:
\begin{itemize}
    \item We propose Ordinal Degradation CLIP (OD-CLIP), which learns a type-aware ordinal geometry for degradation representation. By disentangling degradation type from continuous severity, OD-CLIP distinguishes which degradations are present while the severity ordering within each degradation type, providing an explicit degradation-aware condition for blind SR. 

    \item OD-CLIP organizes degradation representations according to a perceptual severity score, defined as the perceptual deviation induced by each degradation level relative to the clean image. Since physical degradation parameters differ in range and severity direction across degradation types, this score provides a perceptually grounded basis for ordering and spacing severity levels.

    \item We construct type-specific severity anchors by augmenting frozen CLIP type embeddings with learnable level-specific shifts. An ordinal metric loss organizes the anchors according to perceptual severity, while an anchor-guided alignment loss maps visual degradation features onto the resulting ordinal geometry, enabling continuous severity estimation.

    \item We evaluate OD-CLIP as both a representation and a conditioning signal, showing improved severity ranking and perceptual distance modeling over CLIP and existing degradation encoders. When integrated into SeeSR and DiT4SR, OD-CLIP improves fidelity and reduces content inconsistency across unknown and mixed degradations.
\end{itemize}

\section{Related Work}
\label{sec:related_works}

\subsection{Blind Image Super-Resolution}

Blind SR methods can be broadly categorized into degradation-estimation-based approaches and generative-based approaches. Early  works~\cite{ReDegNet, manet, and, bsrdm} explicitly model degradation parameters (e.g., blur kernels and noise levels) and condition the reconstruction network on the estimated degradation. SRMD~\cite{srmd} incorporates degradation vectors into the SR model, while DAN~\cite{dan} alternates between degradation estimation and image reconstruction. Other approaches learn degradation-aware representations jointly with restoration to implicitly capture degradation characteristics. GAN-based blind SR methods further improve perceptual quality through adversarial training, though they may introduce artifacts under challenging degradations.
Recently, diffusion models have been introduced into blind SR, leveraging strong generative priors for high-frequency detail synthesis. Early diffusion-based SR methods~\cite{IDM, DDRM, SR3, DDNM} primarily model the domain shift between downsampled LR and HR images. Subsequent works adapt large-scale text-to-image diffusion models such as Stable Diffusion for restoration. Several approaches~\cite{stablesr, diffbir, pasd, faithdiff} use the low-quality (LQ) image as a visual prior to guide the denoising process. For example, DiffBIR~\cite{diffbir} adopts a two-stage pipeline that first removes degradations and then enhances realism via diffusion, while PASD~\cite{pasd} introduces an auxiliary degradation removal module with multi-scale supervision to extract cleaner features for diffusion control. Prompt-guided methods~\cite{seesr, spire} further incorporate textual descriptions to provide semantic cues during sampling.

\subsection{Degradation Representation of Blind SR}
Since degradation estimation is the main challenge in blind-SR, recent studies \cite{daclip, mmrealsr, supir, spire} have explored explicit modeling and representation of degradation features. DA-CLIP \cite{daclip, wild_clip} separates semantic and degradation embeddings by using diverse text prompts: one branch follows CLIP for semantic alignment, while another learns degradation types (e.g., noise, hazy) via an additional image encoder. MM-RealSR \cite{mmrealsr} performs unsupervised degradation estimation via metric learning by categorizing degradations into noise and blur. ZFusion \cite{zfusion} aligns real-world and synthetic degradation attributes using CLIP embeddings to enhance degradation representation from the LQ image. Although these methods provide new ideas for describing degradation, they often lack explicit definitions or measurable representations of degradation severity. Measuring degradation intensity remains challenging because it involves continuous and interdependent distortions (e.g., noise, blur, JPEG) that are difficult to model within a unified latent space. 
Therefore, we introduce an ordinal text embedding space designed to encode degradation descriptions with explicit ordering. By aligning the image’s degradation embedding with this ordinal textual space, we enforce ordered structural relationships that enhance interpretability and support fine-grained degradation understanding.


\subsection{Numerical Representations using VLM}
The CLIP text encoder exhibits limited sensitivity to numerical variations in prompts. As shown in Fig.~\ref{fig:clip_problem}, different numeric values produce nearly identical embeddings, limiting the representation of quantitative concepts such as degradation levels. Prior studies \cite{countclip, crowdclip, ordinalCLIP, numclip} have pointed out that CLIP’s pretraining data and prompts rarely describe images in terms of quantities, making it inherently weak at understanding numerical relationships.
To address this limitation, CountCLIP \cite{countclip} and CrowdCLIP \cite{crowdclip} introduce numerically descriptive prompts and corresponding images to help CLIP learn object counting. OrdinalCLIP \cite{ordinalCLIP} and NumCLIP \cite{numclip} adopt learning rank-aware text prototypes that enable ordinal regression, allowing the model to represent ordered attributes such as image quality, or human age.
Motivated by this limitation, we reformulate the textual representation space to explicitly encode ordinal relations among degradation levels, rather than relying on CLIP’s lack of an ordinal numerical geometry. By redefining the text embeddings in an ordinal manner, we encourage the image encoder to align with degradation levels and capture their continuous progression, rather than treating them as discrete or unrelated categories.

\begin{figure*}[!t]
    \centering
    \includegraphics[width=1.0\linewidth]{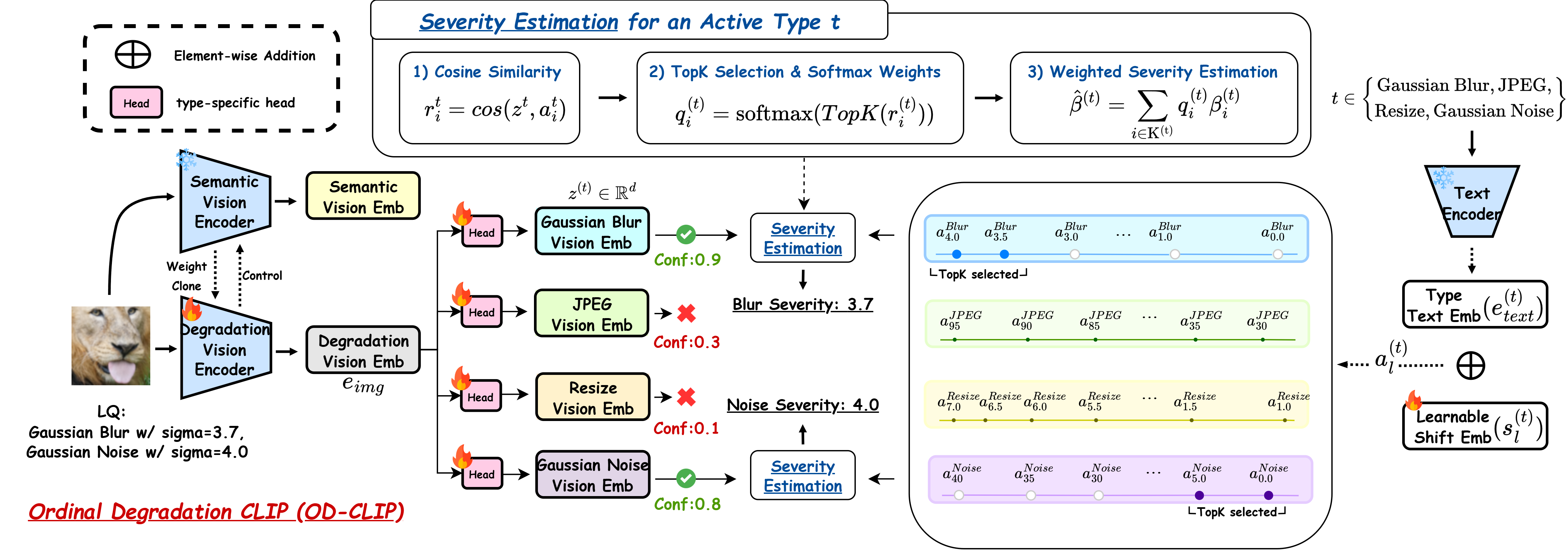}
    \caption{\textbf{Overview of OD-CLIP.} Given an LQ input, we employ a dual encoder to extract semantic and degradation representations. The text encoder produces a type embedding $\mathbf{e}_{\mathrm{text}}^{(t)}$, which is combined with a learnable type- and level-specific shift $\mathbf{s}_{l}^{(t)}$ via element-wise addition to form the severity anchor $\mathbf{a}_{l}^{(t)}$. To infer the degradation state, confidence scores identify the active degradation types, while cosine similarities between the corresponding type-specific visual features and severity anchors measure their compatibility. For each active type, the top-$K$ anchors are selected, and their perceptual severity scores are aggregated using softmax-normalized similarity weights to produce a continuous severity estimate.}
    \label{fig:odclip}

\end{figure*}

\section{Methodology}
\label{sec:method}

\subsection{Problem Setting}
\label{sec:problem_setting}

We first define the degradation setting considered in this work.
Following a classical degradation model commonly adopted in blind image
super-resolution \cite{bsrgan, realesrgan}, a low-quality
image can be expressed as
\begin{equation}
    \mathbf{x}
    =
    \mathcal{D}(\mathbf{y})
    =
    \left[
        \mathcal{R}_{s}
        \left(
            \mathbf{y}\ast\mathbf{k}
        \right)
        +
        \mathbf{n}
    \right]_{\mathrm{JPEG}(q)},
    \label{eq:degradation_model}
\end{equation}
where $\mathbf{y}$ is a clean image, $\mathbf{k}$ is a blur kernel,
$\mathcal{R}_{s}$ denotes downsampling governed by scale factor $s$,
$\mathbf{n}$ denotes additive noise, and $q$ is the JPEG quality factor.

In our controlled setting, we model blur using isotropic Gaussian kernels
and noise as additive Gaussian noise. Accordingly, we consider four
constituent degradation types:
\begin{equation}
\begin{aligned}
    \mathcal{T}=\{&
        \text{Gaussian blur},\ \text{downsampling},\\
        &\text{Gaussian noise},\ \text{JPEG compression}
    \}.
\end{aligned}
\label{eq:degradation_types}
\end{equation}

For \(t \in \mathcal{T}\), let \(l\) denote the corresponding scalar physical degradation parameter, whose definition and range are specified in Sec.~\ref{sec:datasets_and_implementation_details}. A degradation may occur either individually or jointly with other types. Given only the degraded image \(\mathbf{x}\), our objective is to identify which types in \(\mathcal{T}\) are present and estimate the continuous perceptual severity of each present degradation. While this study focuses on these four types, the formulation is not tied to a fixed degradation set and can be extended to include additional types.

\subsection{Ordinal Degradation CLIP}
\label{sec:odclip}

To address the task defined above, we propose Ordinal Degradation CLIP (OD-CLIP), which introduces an explicit ordinal structure into CLIP-based degradation representations. As illustrated in Fig.~\ref{fig:odclip}, OD-CLIP decomposes a shared degradation representation into type-specific visual features and aligns them with type-specific anchors ordered according to perceptual severity.


Given a low-quality image $\mathbf{x}$, the degradation vision encoder
$E_d$ extracts a shared degradation representation:
\begin{equation}
    \mathbf{e}_{\mathrm{img}} = E_d(\mathbf{x}).
\end{equation}
For each degradation type $t\in\mathcal{T}$, a type-specific prediction
head $h_t$ produces a visual feature and a presence probability:
\begin{equation}
    \left(
        \mathbf{z}^{(t)},
        \widehat{p}^{(t)}
    \right)
    =
    h_t
    \left(
        \mathbf{e}_{\mathrm{img}}
    \right),
    \qquad
    t\in\mathcal{T},
    \label{eq:type_specific_head}
\end{equation}
where $\mathbf{z}^{(t)}$ is the visual representation associated with
type $t$, and $\widehat{p}^{(t)}\in[0,1]$ is obtained using a sigmoid
function and indicates the probability that type $t$ is present.
The confidence scores are supervised using an MSE loss, denoted by $\mathcal{L}_{\mathrm{conf}}$. Since each type is predicted independently,
OD-CLIP can identify multiple degradations in the same image.

\subsection{Perceptual Severity Scores}
\label{sec:perceptual_severity_scores}

Across degradation type $t\in\mathcal{T}$, the corresponding physical degradation parameters \(l\) span different numerical ranges and act in opposing directions, and uniformly spaced values of $l$ do not correspond to uniformly spaced perceptual severity.
We therefore introduce a type-specific perceptual severity score $\beta_l^{(t)}$.
Given a clean image $\mathbf{y}$ and its degraded counterpart $D_t(\mathbf{y};l)$, we define:
\begin{equation}
    \beta_l^{(t)}
    =
    \mathbb{E}_{\mathbf{y}\sim\mathcal{D}}
    \left[
        \operatorname{LPIPS}
        \left(
            \mathbf{y},
            D_t(\mathbf{y};l)
        \right)
    \right],
    \label{eq:beta}
\end{equation}
where $\mathcal{D}$ denotes the distribution of clean training images.
This mapping assigns to each physical level $l$ a perceptual severity score $\beta_l^{(t)}$, where a larger score reflects a greater expected perceptual deviation from the clean reference.
These scores form a perceptually spaced severity scale that varies nearly monotonically with degradation strength.

\subsection{Severity Anchors for Each Degradation Type}
\label{sec:ordinal_anchors}

Given the severity score $\beta_l^{(t)}$, we construct a learnable
anchor $\mathbf{a}_l^{(t)}$ for each degradation type $t$ and level $l$:
\begin{equation}
    \mathbf{a}_l^{(t)}
    =
    \mathbf{e}_{\mathrm{text}}^{(t)}
    +
    \mathbf{s}_l^{(t)},
    \label{eq:ordinal_anchor}
\end{equation}
where $\mathbf{e}_{\mathrm{text}}^{(t)}$ is the type embedding produced by
the frozen CLIP text encoder, and $\mathbf{s}_l^{(t)}$ is a learnable
type- and level-specific shift. The type embedding provides a categorical semantic reference, while the shift allows each severity level to occupy a distinct position in the CLIP embedding space.
Our analysis shows that these scores exhibit a consistent monotonic
trend with physical degradation strength over the considered parameter
ranges.


\subsection{Learning the Ordinal Geometry}
\label{sec:ordinal_geometry}
Since association alone does not establish an ordinal structure in the embedding space, we impose it through two complementary objectives. 
An \emph{ordinal metric loss} $\mathcal{L}_{\mathrm{ord}}$ shapes the anchor geometry so that distances between anchors reflect differences in $\beta_l^{(t)}$, and an \emph{anchor-guided alignment loss} $\mathcal{L}_{\mathrm{align}}$ transfers this geometry to the
visual representations.

\vspace{0.1in} \noindent\textbf{Ordinal Anchor Geometry.}
For a pair of anchors ($\mathbf{a}_i^{(t)},\mathbf{a}_j^{(t)}$) from the same degradation type $t$, we define their embedding distance and the difference between their associated severity values as:
\begin{equation}
    D_{ij}^{(t)}
    =
    1-
    \cos
    \left(
        \mathbf{a}_i^{(t)},
        \mathbf{a}_j^{(t)}
    \right),
    \qquad
    B_{ij}^{(t)}
    =
    \left|
        \beta_i^{(t)}-\beta_j^{(t)}
    \right|.
\end{equation}

Let $P$ denote the set of all within-type anchor pairs, and let
$\overline{D}$ and $\overline{B}$ denote their mean embedding distance and mean severity difference, respectively. We define the ordinal metric loss as:
\begin{equation}
    \mathcal{L}_{\mathrm{ord}}
    =
    \frac{1}{|P|}
    \sum_{(t,i,j)\in P}
    \left|
        \frac{D_{ij}^{(t)}}{\overline{D}}
        -
        \frac{B_{ij}^{(t)}}{\overline{B}}
    \right|.
    \label{eq:sm}
\end{equation}

This objective encourages the relative distances between anchors to reflect their corresponding severity differences.

\vspace{0.1in} \noindent\textbf{Anchor-guided visual alignment.}
To transfer the learned anchor geometry to the visual representation, we construct a continuous target for each degradation type $t$. 
Given its ground-truth physical level $l_{GT}^{(t)}$, we first
locate two adjacent sampled levels that bracket it,
$l_k^{(t)} \leq l_{GT}^{(t)} \leq l_{k+1}^{(t)}$.
The corresponding anchors $\mathbf{a}_k^{(t)}$ and
$\mathbf{a}_{k+1}^{(t)}$ are associated with severity coordinates
$\beta_k^{(t)}$ and $\beta_{k+1}^{(t)}$, respectively.
Under the nearly monotonic severity mapping described in Sec.~\ref{sec:perceptual_severity_scores}, these severity values determine the relative position of the target
between the two anchors. We define the interpolation coefficient as:
\begin{equation}
    u
    =
    \frac{
        \beta_{GT}^{(t)}-\beta_k^{(t)}
    }{
        \beta_{k+1}^{(t)}-\beta_k^{(t)}
    },
    \label{eq:interpolation_weight}
\end{equation}
where $u$ = $0$ corresponds to $\mathbf{a}_k^{(t)}$, $u$ = $1$ corresponds to
$\mathbf{a}_{k+1}^{(t)}$, and an intermediate value specifies a position between them.

Given the interpolation coefficient $u$, we obtain an intermediate embedding
target for the GT severity $\beta_{\mathrm{GT}}^{(t)}$ by spherical
interpolation between the two bracketing anchors:
\begin{equation}
    \mathbf{a}^{\star}(\beta_{\mathrm{GT}}^{(t)})
    =
    \frac{
        \sin\!\left((1-u)\theta_k^{(t)}\right)\mathbf{a}_k^{(t)}
        +
        \sin\!\left(u\theta_k^{(t)}\right)\mathbf{a}_{k+1}^{(t)}
    }{
        \sin\theta_k^{(t)}
    },
    \label{eq:slerp_target}
\end{equation}
where $\theta_k^{(t)}$ is the angular distance between the two anchors,
\begin{equation}
    \theta_k^{(t)}
    =
    \arccos
    \left(
        \left\langle
            \mathbf{a}_k^{(t)},
            \mathbf{a}_{k+1}^{(t)}
        \right\rangle
    \right),
    \label{eq:angular_distance}
\end{equation}
and both anchors are $L_2$-normalized beforehand so that the inner product equals their cosine similarity.

We then align the type-specific visual feature with this target by
minimizing their cosine distance:
\begin{equation}
    \mathcal{L}_{\mathrm{align}}^{(t)}
    =
    1-
    \cos
    \left(
        \mathbf{z}^{(t)},
        \mathbf{a}^{\star}(\beta_{\mathrm{GT}}^{(t)})
    \right).
    \label{eq:alignment_loss}
\end{equation}

Together, $\mathcal{L}_{\mathrm{ord}}$ structures the anchor space according to severity, while $\mathcal{L}_{\mathrm{align}}$ matches the visual features with their corresponding positions in this space.

\subsection{Severity Estimation}
\label{sec:severity_estimation}
To estimate the severity of degradation type $t$, we compare
$\mathbf{z}^{(t)}$ with the severity anchors defined for that type. The
similarity score for anchor $i$ is:
\begin{equation}
    r_i^{(t)}
    =
    \cos
    \left(
        \mathbf{z}^{(t)},
        \mathbf{a}_i^{(t)}
    \right),
    \label{eq:anchor_similarity}
\end{equation}
then select the $k$ anchors with
the highest similarity scores.
\begin{equation}
    \mathcal{K}^{(t)}
    =
    \operatorname{TopK}
    \left(
        \left\{
            r_i^{(t)}
        \right\}_{i=1}^{N_t},
        k
    \right),
    \label{eq:topk_selection}
\end{equation}
where $N_t$ is the number of anchors associated with degradation type $t$.
The selected similarity scores are normalized using a softmax to obtain
their weights:
\begin{equation}
    q_i^{(t)}
    =
    \frac{
        \exp\left(r_i^{(t)}\right)
    }{
        \sum_{j\in\mathcal{K}^{(t)}}
        \exp\left(r_j^{(t)}\right)
    },
    \qquad
    i\in\mathcal{K}^{(t)},
    \label{eq:topk_probability}
\end{equation}
Since the softmax
is computed over the selected anchors only, $\sum_{i\in\mathcal{K}^{(t)}} q_i^{(t)} = 1$.
Using these weights, we estimate the continuous severity as the weighted
average of their corresponding perceptual severity scores:
\begin{equation}
    \widehat{\beta}^{(t)}
    =
    \sum_{i\in\mathcal{K}^{(t)}}
    q_i^{(t)}
    \beta_i^{(t)}.
    \label{eq:beta_readout}
\end{equation}
For each ground-truth active type, the severity estimate is supervised using
an $\ell_1$ loss:
\begin{equation}
    \mathcal{L}_{\mathrm{level}}^{(t)}
    =
    \left|
        \widehat{\beta}^{(t)}
        -
        \beta_{\mathrm{GT}}^{(t)}
    \right|.
    \label{eq:severity_loss}
\end{equation}

\vspace{0.1in} \noindent\textbf{Overall objective.}
Our objective jointly preserves semantic content and learns the ordinal
degradation representation:
\begin{equation}
\begin{aligned}
    \mathcal{L}_{\mathrm{total}}
    ={}&
    \mathcal{L}_{\mathrm{CLIP}}
    +
    \mathcal{L}_{\mathrm{GT}}
    +
    \mathcal{L}_{\mathrm{conf}}
    \\
    &+
    \mathcal{L}_{\mathrm{ord}}
    +
    \mathcal{L}_{\mathrm{align}}
    +
    \mathcal{L}_{\mathrm{level}}.
\end{aligned}
\end{equation}
where $\mathcal{L}_{\mathrm{CLIP}}$ and $\mathcal{L}_{\mathrm{GT}}$ preserve semantic consistency following DA-CLIP~\cite{daclip, wild_clip}, while
the remaining terms supervise degradation presence, anchor geometry, visual alignment, and severity estimation, respectively. (Details are provided in the supplementary material).

\subsection{Conditioning with OD-CLIP for Blind SR}
To assess the practical utility of the learned degradation representations, we incorporate OD-CLIP into two diffusion-based blind SR frameworks, SeeSR~\cite{seesr} and DiT4SR~\cite{dit4sr}. As these frameworks differ in their backbone architectures and conditioning mechanisms, we adopt a framework-specific integration strategy for each model rather than imposing a unified design. We adapt the components responsible for processing conditional information, such as cross-attention or modulation layers, to accommodate OD-CLIP representations during restoration. This enables the restoration models to leverage information about both degradation type and severity. We evaluate OD-CLIP under different training settings, including model retraining and the fine-tuning of a pre-trained restoration model. The results show that incorporating OD-CLIP improves restoration fidelity, highlighting the practical value of ordinal degradation representations for guiding diffusion-based blind SR. Further implementation and training details are provided in the supplementary material.

\section{Experimental Results}
\label{sec:experimental_results}
\subsection{Datasets and implementation details}
\label{sec:datasets_and_implementation_details}

\noindent\textbf{Training Datasets.}
We train OD-CLIP on DIV2K training set~\cite{div2k}. Our framework requires paired HR–LR data with known degradation parameters. Following SPIRE~\cite{spire}, we synthesize the four degradation types defined in Sec.~\ref{sec:problem_setting}: isotropic Gaussian blur with $\sigma_b\in[0.1,4.0]$, additive Gaussian noise with $\sigma_n\in[1,40]$, JPEG compression with quality factor $q\in[30,95]$, and downsampling with scale factor $s\in[1.1,7.0]$.

For each DIV2K training image, we generate multiple degraded versions with continuously varying intensities. This multi-level sampling enables OD-CLIP to distinguish degradation intensity from image content. We also construct composite samples by sequentially applying multiple degradation types with randomized intensities, encouraging robust representations that generalize to complex degradations.

\vspace{0.1in} \noindent\textbf{Test datasets and Evaluation Metrics.}
We evaluate OD-CLIP on a validation set synthesized from DIV2K-Val~\cite{div2k} using the same degradation pipeline. Type Accuracy measures whether each degradation type is correctly
identified as present or absent, using $\hat{p}^{(t)}$>$0.5$ to indicate presence and averaging the accuracy across the four types. MAE, SROCC, and PCC are evaluated in the perceptual $\beta$-space over correctly detected active types and averaged across types; MAE is normalized by $\beta_{\max}^{(t)}$.

To validate the effectiveness of OD-CLIP embeddings when integrated into pre-trained SR models, we evaluate on DIV2K-Val with degradations synthesized by the Real-ESRGAN pipeline~\cite{realesrgan}, and on two real-world benchmarks, DRealSR~\cite{drealsr} and RealSR~\cite{realsr}. 
Performance is reported with reference-based fidelity metrics (PSNR and SSIM), perceptual metrics (LPIPS~\cite{lpips} and DISTS~\cite{dists}), and FID~\cite{fid} to measure distribution-level alignment with the ground truth. 
This protocol covers both synthesized and real-world degradations, and evaluates structural fidelity alongside perceptual realism.
Throughout all tables, the best and second-best results are highlighted in \best{Red} and \second{Blue}, respectively.

\vspace{0.1in} \noindent\textbf{Implementation details.}
Our framework is implemented in two stages.
In the first stage, we train the proposed OD-CLIP model based on the CLIP backbone (ViT-B/32) using a two-phase training schedule.
We begin by training on single-degradation data for 50 epochs to establish the per-type anchor curves, then continue training for another 150 epochs on multi-degradation data to adapt the model to compositional inputs.
Severity anchors are uniformly sampled over their physical degradation parameter range $(\Delta l{=}1$), yielding $N_t$ anchors, and the $\operatorname{TopK}=2$ are used for severity estimation. Training is conducted on four NVIDIA RTX 3090 GPUs with a total batch size of $2{,}048$, using the AdamW optimizer with a learning rate of $2\times10^{-4}$. The entire training procedure takes approximately five hours.
In the second stage, the pre-trained OD-CLIP is integrated into the SR frameworks using framework-specific adaptations. Both training and evaluation for the baseline and OD-CLIP-enhanced models are conducted on a single NVIDIA RTX Pro 6000 GPU.

\begin{table}[t]
    \centering    
    \small
    \setlength{\tabcolsep}{1mm}
    \begin{tabular}{lcccc}
    \toprule
    Method & Type Acc $\uparrow$ & $\mathrm{MAE}$ $\downarrow$ & SROCC $\uparrow$ & PCC $\uparrow$ \\
    \midrule
    CLIP
        & 58.6\%
        & 0.328
        & -0.137
        & -0.113 \\
    
    DA-CLIP
        & 69.2\%
        & 0.257
        & -0.010
        & 0.005 \\
    
    \textbf{OD-CLIP (ours)}
        & \best{84.8\%}
        & \best{0.122}
        & \best{0.888}
        & \best{0.867} \\
    \bottomrule
    \end{tabular}
    \caption{ Comparison of degradation representation.}
    \label{tab:model_comparison}
\end{table}

\begin{figure}[!t]
    \centering
    \begin{subfigure}[b]{0.48\linewidth}
        \centering
        \includegraphics[width=0.8\linewidth]{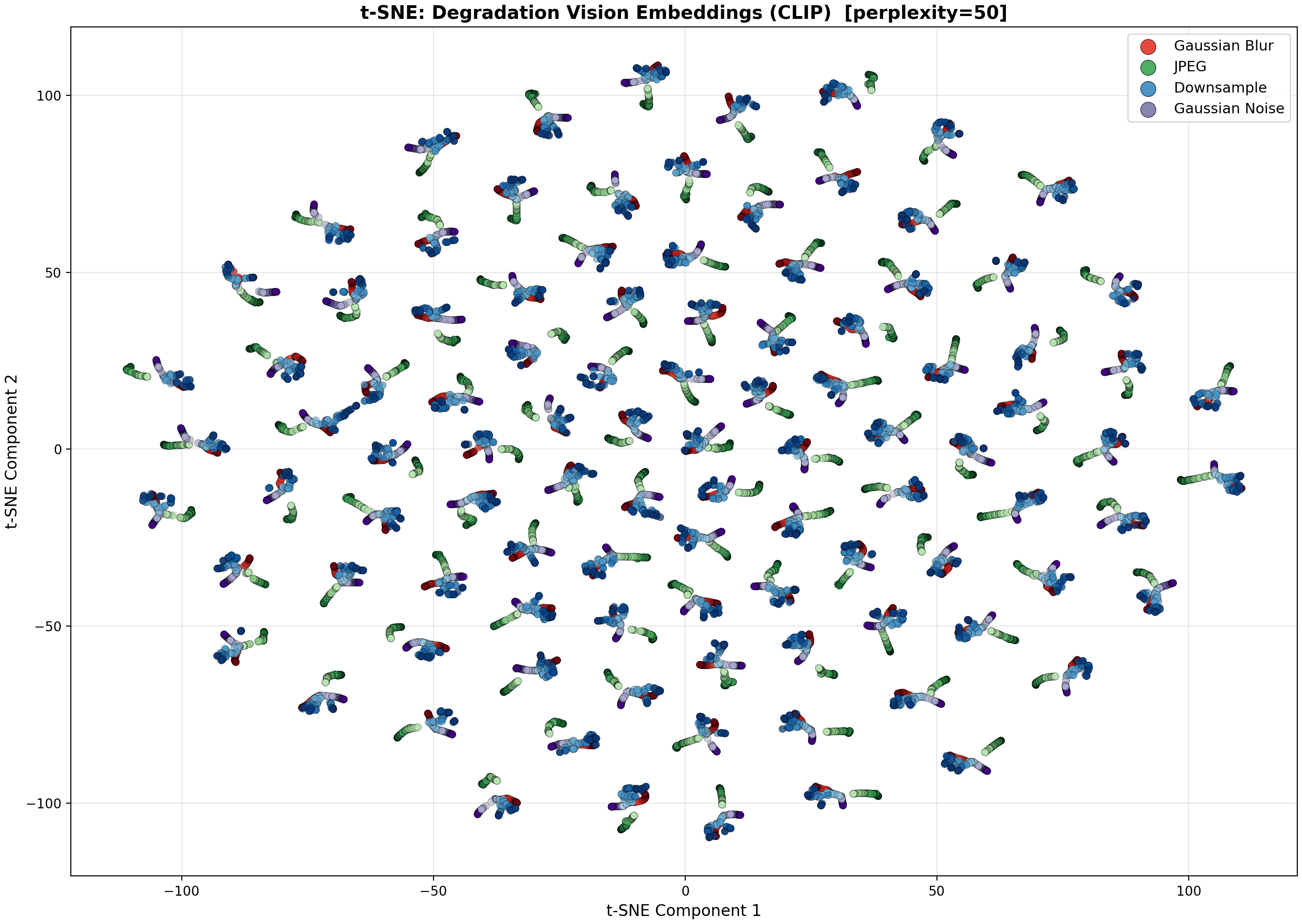}
        \caption{CLIP}
        \label{fig:tsne_clip}
    \end{subfigure}
    \hfill
    \begin{subfigure}[b]{0.48\linewidth}
        \centering
        \includegraphics[width=0.8\linewidth]{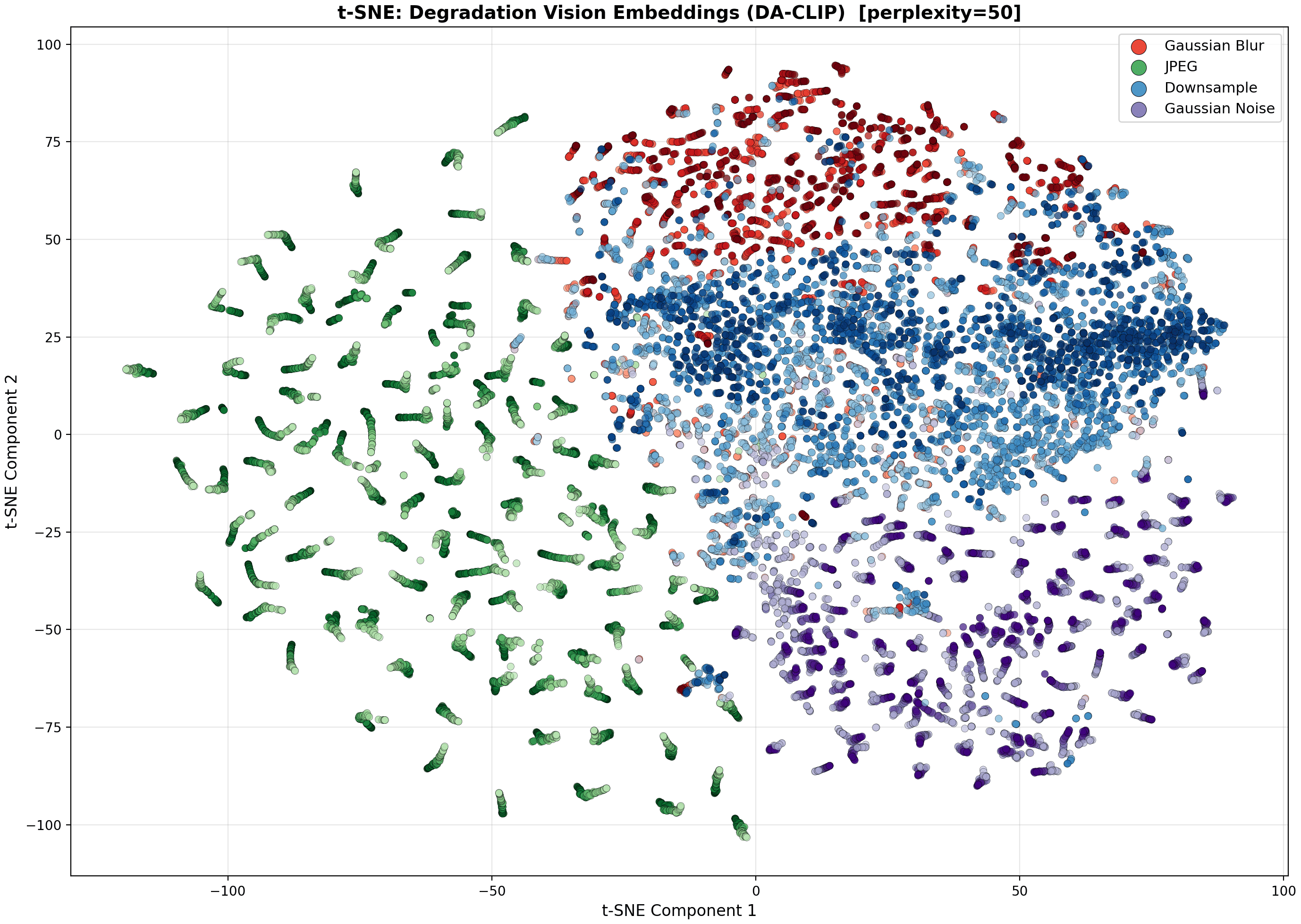}
        \caption{DA-CLIP}
        \label{fig:tsne_daclip}
    \end{subfigure}
    
    \vspace{1em} 
    
    \begin{subfigure}[b]{0.48\linewidth}
        \centering
        \includegraphics[width=0.8\linewidth]{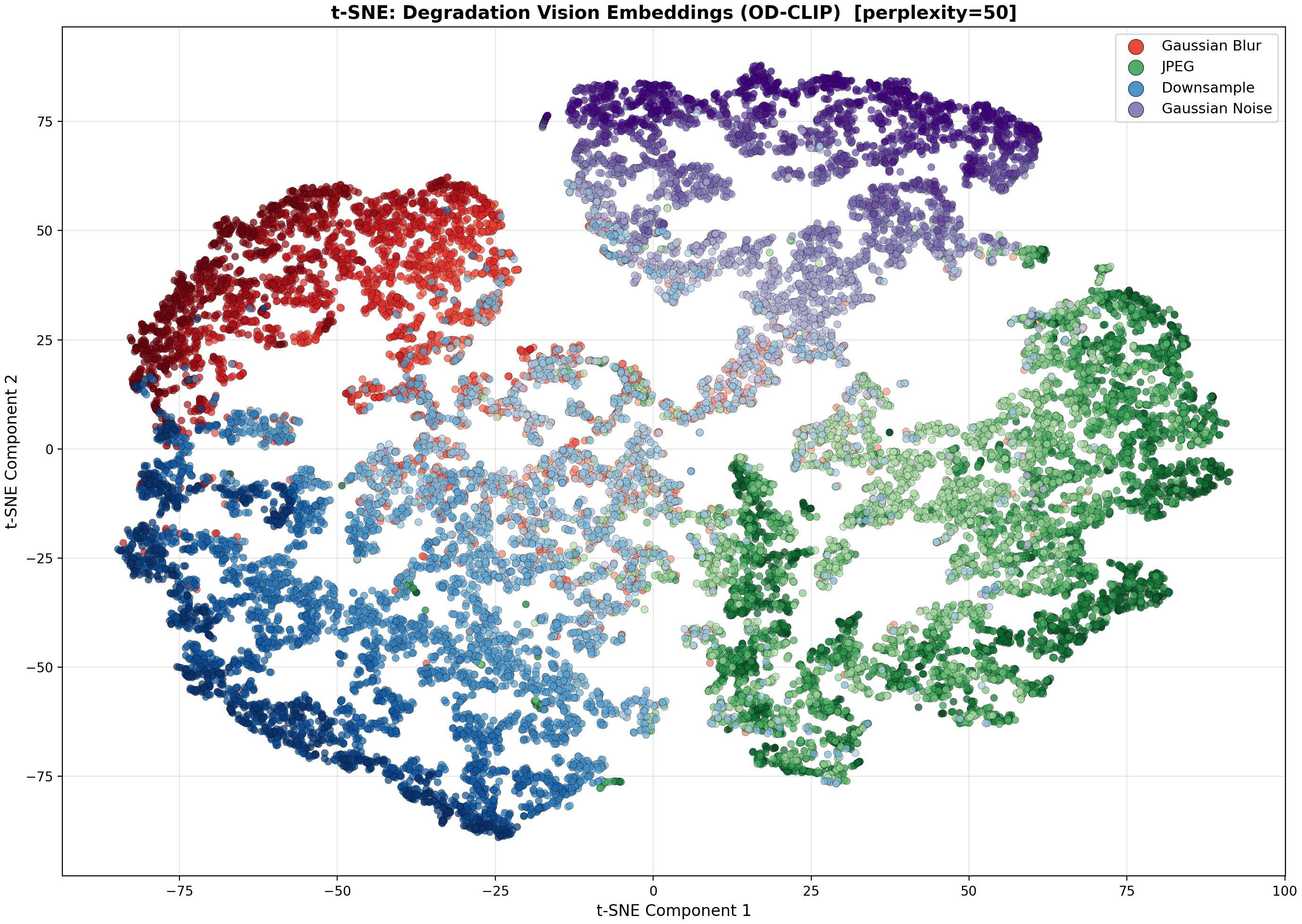}
        \caption{OD-CLIP (Ours)}
        \label{fig:tsne_odclip_final}
    \end{subfigure}
    \hfill
    \begin{subfigure}[b]{0.48\linewidth}
        \centering
        \includegraphics[width=0.8\linewidth]{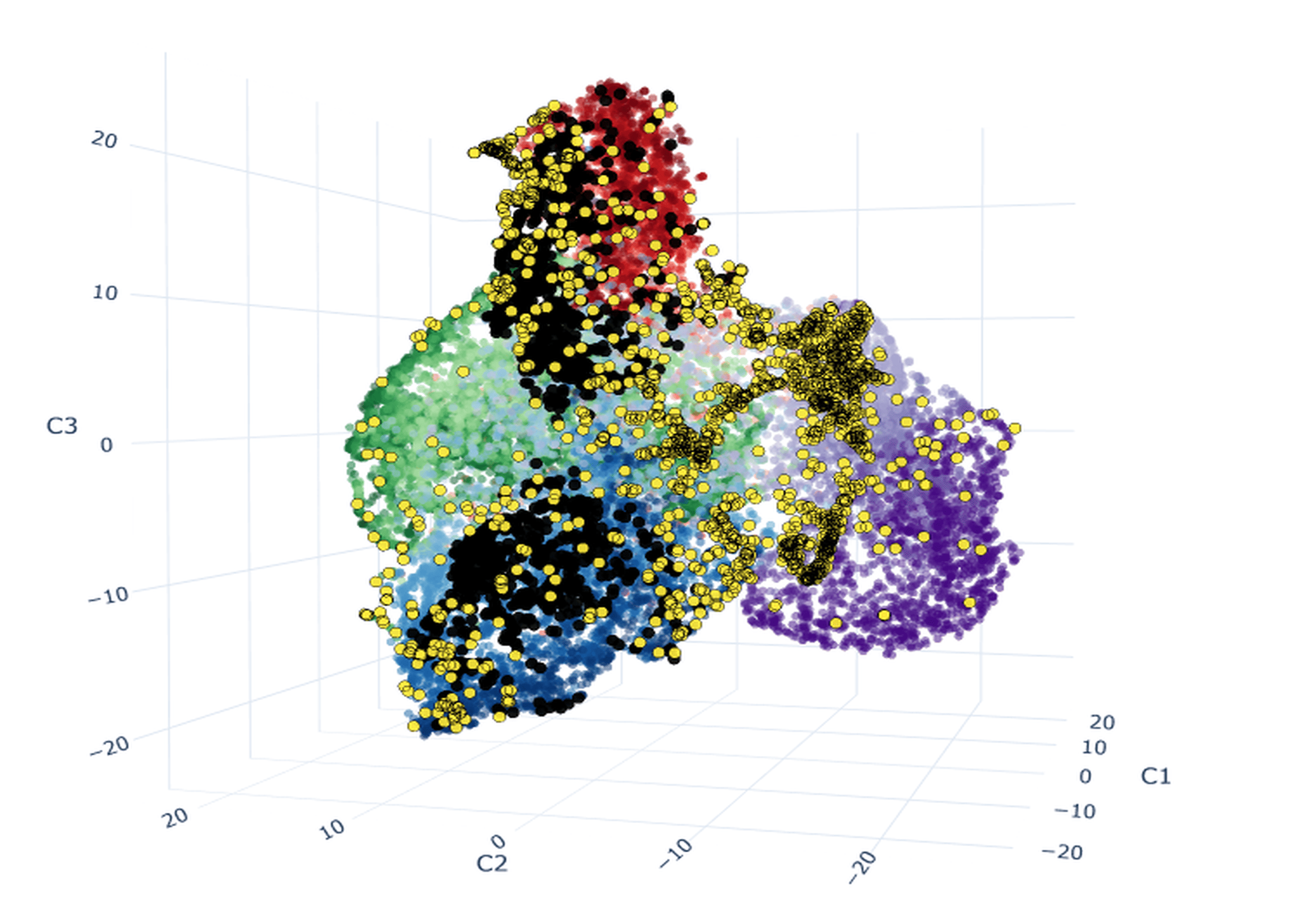}
        \caption{3D t-SNE of OD-CLIP}
        \label{fig:tsne_odclip_variant}
    \end{subfigure}
    
    \caption{t-SNE visualization of degradation embeddings. (a)–(c) CLIP, DA-CLIP, and OD-CLIP embeddings across degradation types and severity levels. OD-CLIP produces well-separated clusters with ordered intensity structures. (d) 3D t-SNE of OD-CLIP embeddings for OOD degradations synthesized by Real-ESRGAN \cite{realesrgan} (black) and BSRGAN \cite{bsrgan} (yellow). Their integration into the known degradation manifold indicates robust generalization to complex and mixed artifacts.}
    \label{fig:tsne_ODCLIP}
\end{figure}

\subsection{Comparison of OD-CLIP for Degradation Severity Estimation}
We evaluate the representational capability of OD-CLIP against vanilla CLIP~\cite{clip} and DA-CLIP~\cite{daclip, wild_clip} in Table~\ref{tab:model_comparison}. CLIP shows limited degradation awareness because its text encoder, optimized for semantic concepts, cannot effectively align visual features with degradation representations. DA-CLIP improves degradation-type discrimination through supervised fine-tuning, but treats different intensities as categorical labels, limiting its ability to capture fine-grained severity differences. In contrast, OD-CLIP models the ordinal topology of degradation through level-wise embeddings, ensuring that the latent space reflects the monotonic progression of intensity. Learnable shift embeddings further capture perceptual variations across levels, producing a manifold aligned with both degradation category and severity.
This capability is qualitatively validated by the t-SNE visualizations in Fig.~\ref{fig:tsne_ODCLIP} (a)--(c). CLIP fails to form distinct clusters, confirming its limited degradation awareness. DA-CLIP achieves partial category separation but still exhibits intra-class overlap across intensities. In contrast, OD-CLIP produces well-separated clusters that preserve severity gradients, demonstrating that its representations capture both degradation semantics and intensity.

Finally, Fig.~\ref{fig:tsne_ODCLIP} (d) uses 3D t-SNE to evaluate the generalization of OD-CLIP to unseen degradation regimes. We visualize representations of images synthesized by Real-ESRGAN~\cite{realesrgan} (black points) and BSRGAN~\cite{bsrgan} (yellow points), which contain OOD degradation factors. These samples lie within the manifold of known degradations, indicating robustness to complex, stochastic, and mixed artifacts. This suggests that OD-CLIP can help diffusion models better interpret and correct complex real-world degradations during restoration.

\begin{table}[t]
\centering
    \small
    \setlength{\tabcolsep}{1mm}
    \begin{tabular}{l ccccc}
    \toprule
    & \multicolumn{2}{c}{\emph{Fidelity}} & \multicolumn{2}{c}{\emph{Perceptual}} &  \emph{Distribution}\\
    \cmidrule(lr){2-3}  \cmidrule(lr){4-5} \cmidrule(lr){6-6}
    \textbf{Method} & \textbf{PSNR}$\uparrow$ & \textbf{SSIM}$\uparrow$ & \textbf{LPIPS}$\downarrow$ & \textbf{DISTS}$\downarrow$ & \textbf{FID}$\downarrow$ \\
    \midrule
    \multicolumn{6}{c}{\textbf{DIV2K-Val} \cite{div2k}} \\
    \midrule
    SUPIR & 23.1629 & 0.5445 & 0.3631 & 0.2259 & 28.0102\\
    DiffBIR & 23.1450 & 0.5441 & 0.3669 & 0.2209 & 32.7289 \\
    XPSR & 22.7468 & 0.5656 & 0.3638 & 0.2202 & 30.4971\\
    FaithDiff & 23.4899 & 0.5812 & 0.3118 & 0.1988 & 25.6303 \\
    VOSR & \second{23.8166} & 0.6028 & \best{0.3041} & \second{0.1947} & \best{24.1065} \\
    \midrule
    DiT4SR & 22.0289 & 0.5574 & 0.3417 & 0.2120 & 36.4627 \\
    + ODCLIP & 22.0962 & 0.5481 & 0.3377 & 0.2013 & 26.8002 \\
    SeeSR & 23.7250 & \second{0.6056} & 0.3194 & 0.1966 & 25.8199\\
    + ODCLIP & \best{24.4334} & \best{0.6321} & \second{0.3096} & \best{0.1912} & \second{24.8198}\\
    \midrule
    \multicolumn{6}{c}{\textbf{DRealSR} \cite{drealsr}} \\
    \midrule
    SUPIR & 27.0085 & 0.6709 & 0.4262 & 0.2761 & \second{147.4735} \\
    DiffBIR & 25.9019 & 0.6244 & 0.4669 & 0.2882 & 180.6325 \\
    XPSR & 26.1980 & 0.7227 & 0.3814 & 0.2651 & 170.1793 \\
    FaithDiff & 27.3576 & 0.7129 & 0.3505 & 0.2377 & 150.9194 \\
    VOSR & 27.6146 & 0.7463 & 0.3606 & 0.2467 & 175.8697 \\
    \midrule
    DiT4SR & 25.6984 & 0.6775 & 0.3741 & 0.2467 & 158.8172 \\
    + ODCLIP & 26.4681 & 0.6973 & 0.3644 & 0.2353 & 148.1334 \\
    SeeSR & \second{28.1430} & \second{0.7711} & \second{0.3142} & 0.2299 & \best{146.9934} \\
    + ODCLIP & \best{28.8697} & \best{0.8009} & \best{0.2963} & \best{0.2227} & 148.4187 \\
    \midrule
    \multicolumn{6}{c}{\textbf{RealSR} \cite{realsr}} \\
    \midrule
    SUPIR & 25.2268 & 0.6757 & 0.3720 & 0.2479 & 120.6293 \\
    DiffBIR & 24.8342 & 0.6502 & 0.3649 & 0.2399 & 130.7316 \\
    XPSR & 23.5930 & 0.6698 & 0.3646 & 0.2574 & 140.3304 \\
    FaithDiff & 25.2781 & 0.7060 & 0.2885 & \second{0.2108} & \best{111.7713} \\
    VOSR & \second{25.3693} & 0.7195 & \second{0.2867} & \best{0.2086} & 123.8297 \\
    \midrule
    DiT4SR & 23.3994 & 0.6622 & 0.3229 & 0.2238 & 120.3651 \\
    + ODCLIP & 24.0207 & 0.6762 & 0.3135 & 0.2196 & \second{112.9471} \\
    SeeSR & 25.2082 & \second{0.7215} & 0.3004 & 0.2218 & 124.9782 \\
    + ODCLIP & \best{25.8823} & \best{0.7470} & \best{0.2860} & 0.2206 & 135.3407 \\
    \bottomrule
    \end{tabular}
    \caption{Quantitative evaluation on DIV2K-Val, DRealSR, and RealSR benchmarks.
    }
    \label{tab:sr_comparison}
\end{table}

\subsection{Comparisons of Conditioning on Blind-SR}
\label{subsec:adj_CFPG}
\begin{figure}[!t]
    \centering
    \includegraphics[width=1.0\linewidth]{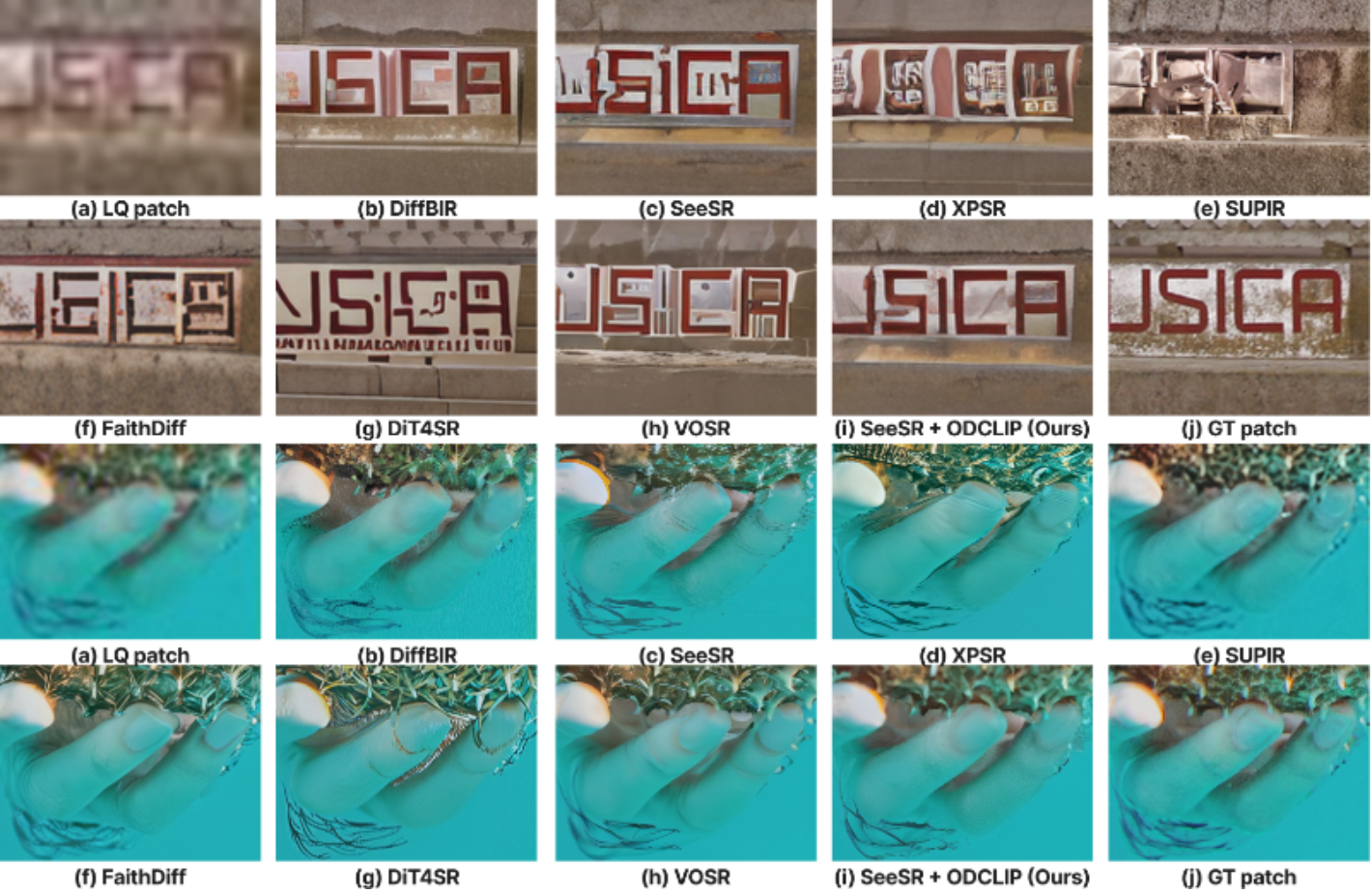}
    \caption{
    Qualitative comparisons with SOTA methods. With OD-CLIP, our method achieves the better fidelity and content structure. Additional visual results are provided in the supplementary material. \textbf{(Zoom in for a better view.)}
    }
    \label{fig:stage2_visualize}
\end{figure}

As shown in Table~\ref{tab:sr_comparison} and Fig.~\ref{fig:stage2_visualize}, our framework generally improves fidelity while maintaining perceptual quality across SD-based SR methods, showing enhanced degradation understanding that strengthens fidelity. These results confirm that OD-CLIP embeddings serve as powerful conditional priors, guiding the diffusion process toward semantically consistent and perceptually faithful restoration.

\begin{table}[t]
    \centering
    \small
    \setlength{\tabcolsep}{1mm}
    \begin{tabular}{c ccc | cccc}
        \toprule
        & $\mathcal{L}_{\mathrm{ord}}$
        & $\mathcal{L}_{\mathrm{align}}$
        & $\mathcal{L}_{\mathrm{level}}$
        & Type Acc$\uparrow$
        & MAE$\downarrow$
        & SROCC$\uparrow$
        & PCC$\uparrow$ \\
        \midrule
        (a) & \checkmark & \checkmark & \checkmark
            & \best{84.8\%} & 0.122
            & \best{0.888} & \best{0.867} \\

        (b) &            & \checkmark & \checkmark
            & 80.4\% & 0.293
            & 0.425 & 0.343 \\

        (c) & \checkmark &            & \checkmark
            & 84.1\% & \best{0.102}
            & 0.840 & 0.831 \\

        (d) & \checkmark & \checkmark &
            & 79.7\% & 0.182
            & 0.766 & 0.704 \\
        \bottomrule
    \end{tabular}
    \caption{Ablation on $\mathcal{L}_{ord}$, $\mathcal{L}_{align}$ and $\mathcal{L}_{level}$. (a–d) denote different loss combinations, all including $\mathcal{L}_{conf}$.}
    \label{tab:loss_ablation}
\end{table}

\section{Ablation Study}
\label{sec:ablation_study}

As shown in Table~\ref{tab:loss_ablation}, the three objectives serve complementary roles. The full model in (a) achieves the best overall balance among classification, regression, and ordering performance. Removing \(L_{\mathrm{ord}}\) in (b) substantially weakens the alignment between predicted \(\beta\) values and perceptual severity, demonstrating its importance in preserving intra-type ordering. Removing \(L_{\mathrm{align}}\) in (c) slightly improves point-wise \(\beta\)-regression accuracy but degrades classification and ordering consistency, indicating that it acts as a geometric regularizer that promotes a smooth and globally consistent anchor manifold. Removing \(L_{\mathrm{level}}\) in (d) leads to broad performance degradation, confirming that direct per-image supervision is important for severity estimation and also supports classification through the shared branch backbone.

\section{Conclusion}
\label{sec:conclusion}

In this work, we addressed the ambiguity of degradation representation in blind SR by proposing OD-CLIP, an ordinal vision-language representation framework that jointly models degradation types and continuous perceptual severity. Unlike existing approaches that treat degradation levels as discrete categories or rely on numerically insensitive textual prompts, OD-CLIP constructs type-specific severity anchors and organizes them according to perceptual degradation distances. By aligning degraded visual features with this ordinal embedding space, OD-CLIP enables fine-grained degradation recognition and continuous severity estimation under both individual and mixed degradation settings.
Integrating OD-CLIP into diffusion-based backbones through fine-tuning substantially improves restoration fidelity. Our experiments demonstrate that the proposed framework yields high-fidelity restoration while supporting interpretable super-resolution under unknown and mixed degradations.

\section*{Acknowledgments}

This work was financially supported in part (project number: 112UA10019) by the Co-creation Platform of the Industry Academia Innovation School, NYCU, under the framework of the National Key Fields Industry-University Cooperation and Skilled Personnel Training Act, from the Ministry of Education (MOE) and industry partners in Taiwan. It also supported in part by the National Science and Technology Council, Taiwan, under Grant NSTC-115-2634-F-A49-011-, NSTC-114-2218-E-A49-024-, Grant NSTC-115-2425-H-A49-001, Grant NSTC-114-2622-E-A49-027, Grant NSTC-115-2221-E-A49 -124 -MY3, Grant NSTC-115-2218-E-A49 -017 and in part by the Higher Education Sprout Project of the National Yang Ming Chiao Tung University and the Ministry of Education (MOE), Taiwan. It is also partly supported by MediaTek Inc., Hon Hai Research Institute, and Industrial Technology Research Institute. (Corresponding author: Ching-Chun Huang.)

\bibliography{aaai2027}

\clearpage
\setcounter{page}{1}
\twocolumn[
\begin{center}
    {\LARGE\bfseries Learning Ordinal Degradation Representations with Textual Priors for Diffusion-Based Blind Image Super-Resolution - Technical Appendices\par}

    \vspace{1.5em}
\end{center}
]

\appendix
\setcounter{figure}{5}
\setcounter{table}{4}
\setcounter{equation}{18}

\section{Additional OD-CLIP Methodology Details}
This section provides the details omitted from the main paper: the anchor grid and its resolution (Sec.~\ref{sec.pdpg_and_anchor_const}), how the perceptual severity scale is built (Sec.~\ref{sec:severity_scale}), and the complete training objective (Sec.~\ref{sec.training_objective}).

\subsection{Physical Degradation Parameter Grid and Anchor Construction}
\label{sec.pdpg_and_anchor_const}

For each degradation type $t$, we first enumerate its physical parameter range using a type-specific interval $\delta_t$, yielding an ordered set of physical degradation levels. We define $\Delta l$ as the index stride used to select anchors from these levels, rather than as a direct increment of the physical parameter. For example, the Gaussian-blur range $[0,4.0]$ is enumerated with $\delta_{\mathrm{blur}}=0.1$ as $\{0.0,0.1,0.2,\ldots,3.8,3.9,4.0\}$. Hence, $\Delta l=1$ selects every enumerated level and produces $N_{\mathrm{blur}}=41$ anchors. The type-specific configurations are reported in Table~\ref{tab:grid}. In Sec.~E, we study coarser anchor sampling by varying the index stride over $\Delta l\in\{1,5,10,20\}$.

\begin{table}[t]
\centering
\setlength{\tabcolsep}{1mm}
\begin{tabular}{lccc}
\toprule
Type & Parameter range & $\delta_t$ & $N_t$ \\
\midrule

Gaussian blur& $\sigma_b \in [0.0,4.0]$& $0.1$& $41$ \\
JPEG & $q \in [30,95]$& $1$& $66$ \\
Downsampling& $s \in [1.0,7.0]$& $0.1$& $61$ \\
Gaussian noise& $\sigma_n \in [0,40]$& $1$& $41$ \\

\bottomrule
\end{tabular}
\caption{
Anchor configurations for the $\Delta l=1$ setting used throughout the main paper. For each degradation type $t$, $\delta_t$ denotes the interval between enumerated physical levels, and $N_t$ denotes the resulting number of anchors.
}
\label{tab:grid}
\end{table}

\begin{table}[t]
\centering
\setlength{\tabcolsep}{1mm}
\begin{tabular}{lccc}
\toprule
Type & $\beta_{\min}$ & $\beta_{\max}$ & Spearman $\rho$ \\
\midrule
Gaussian blur& 0.0000 & 0.6581 & 1.0000 \\

JPEG & 0.0079 & 0.1303 & 1.0000 \\

Downsampling& 0.0000 & 0.5973 & 0.9987 \\

Gaussian noise& 0.0000 & 0.7349 & 1.0000 \\

\bottomrule
\end{tabular}
\caption{
Ranges and ordinal consistency of the precomputed perceptual severity scores. The values are retained in their original, unnormalized LPIPS scale. Spearman's $\rho$ is computed between the physical levels ordered from weaker to stronger degradation and their corresponding severity scores.
}
\label{tab:severity_range}
\end{table}

\begin{table}[t]
\centering
\setlength{\tabcolsep}{1mm}
\begin{tabular}{lcccc}
\toprule
Type &
$25\%$ &
$50\%$ &
$75\%$ &
Severity at midpoint \\
\midrule
Gaussian blur& $0.8$& $1.3$& $2.2$& $0.72$ \\

JPEG & $77$& $59$& $42$& $0.46$ \\

Downsampling& $1.9$& $2.8$& $4.3$& $0.72$ \\

Gaussian noise& $11$& $20$& $29$& $0.51$ \\
\bottomrule
\end{tabular}
\caption{
Nonlinearity of the mapping from physical degradation level $l$ to perceptual severity $\beta_l^{(t)}$. The $25\%$, $50\%$, and $75\%$ columns report the physical parameter values at which the normalized severity reaches the corresponding fraction of its type-specific range. The physical parameters are $\sigma_b$, $q$, $s$, and $\sigma_n$ for Gaussian blur, JPEG compression, downsampling, and Gaussian noise, respectively. The final column reports the normalized severity evaluated at the midpoint of each physical parameter range.
}
\label{tab:severity_nonlinearity}
\end{table}

\begin{algorithm*}[t]
  \caption{OD-CLIP training objective}
  \label{alg:odclip_training}
  \begin{algorithmic}[1]
  \REQUIRE $x$: Degraded image, $y$: Clean image, $c$: Caption 
  
  \REQUIRE $\rm Level^{(t)} =\{l_1^{(t)},\ldots,l_{N_t}^{(t)}\}$: the physical levels selected for constructing severity anchors. Each level $l_i^{(t)}$ is mapped by Eq.~(5) to a perceptual severity coordinate $\beta_i^{(t)}\triangleq\beta_{l_i^{(t)}}^{(t)}$, with which a learnable anchor $a_i^{(t)}$ is associated.
  \REQUIRE $p^{(t)}\in\{0,1\}$: whether degradation type $t$ is present in the input image $x$.
  \REQUIRE $l_{\mathrm{GT}}^{(t)}$: ground-truth physical degradation level in $x$ with $p^{(t)}=1$, and $\beta_{\mathrm{GT}}^{(t)} \triangleq \beta_{l_{\mathrm{GT}}^{(t)}}^{(t)}$ denotes the corresponding perceptual severity score.
  \REQUIRE $E_{\mathrm{sem}}, E_{\mathrm{txt}}$: Frozen CLIP encoders;
           \ $E_d$: Degradation vision encoder, $h_t$: Type-specific prediction head
  \ENSURE  $\mathcal{L}_{\mathrm{total}}$ \hfill (Eq.~18)
  \STATE $\mathbf{e}_{\mathrm{img}} \gets E_d(x)$ 
  \STATE $\mathbf{e}_{\mathrm{GT}} \gets E_{\mathrm{sem}}(y)$, $\mathbf{e}_{\mathrm{sem}} \gets E_{\mathrm{sem}}(x)$, with the hidden states
         of $E_d(x)$ injected into its layers
  \STATE $\mathcal{L}_{\mathrm{CLIP}} \gets
         \mathrm{InfoNCE}(\mathbf{e}_{\mathrm{sem}}, E_{\mathrm{txt}}(c))$, \quad
         $\mathcal{L}_{\mathrm{GT}} \gets
         \|\mathbf{e}_{\mathrm{sem}} - \mathbf{e}_{\mathrm{GT}}\|_1$
  \STATE $(\mathbf{z}^{(t)}, \hat{p}^{(t)}) \gets h_t(\mathbf{e}_{\mathrm{img}})$,
         \ $t \in \mathcal{T}$
  \STATE $\mathcal{L}_{\mathrm{conf}} \gets \frac{1}{|\mathcal{T}|}
         \sum_{t \in \mathcal{T}} \big( \hat{p}^{(t)} - p^{(t)} \big)^2$
  \STATE $e_{\rm text}^{(t)} \gets E_{\rm txt}(\operatorname{prompt}(t)),\quad t\in\mathcal T$
  \STATE $a_i^{(t)} \gets e_{\rm text}^{(t)}+s_i^{(t)}, \quad t\in\mathcal T,\ i=1,\ldots,N_t$ \hfill (Eq.~6)
  \STATE $D^{(t)}_{ij}, B^{(t)}_{ij} \gets$ Eq.~(7);
         \quad $\mathcal{L}_{\mathrm{ord}} \gets$ Eq.~(8) \hfill \emph{within-type anchors only}
         
\FOR{each $t$ with $p^{(t)} = 1$}
    \STATE $r^{(t)}_i \gets \cos(\mathbf{z}^{(t)}, \mathbf{a}^{(t)}_{l_i})$,
           \ $i = 1, \dots, N_t$ \hfill (Eq.~13)
    \STATE $\mathrm{K}^{(t)} \gets \mathrm{TopK}(r^{(t)}, 2)$, \quad
           $q^{(t)} \gets \mathrm{softmax}\big(r^{(t)}_{\mathrm{K}^{(t)}}\big)$
           \hfill (Eqs.~14,\,15)
    \STATE $\hat{\beta}^{(t)} \gets
           \sum_{i \in \mathrm{K}^{(t)}} q^{(t)}_i \, \beta^{(t)}_{l_i}$ \hfil
    \STATE Find adjacent anchor levels $l_k^{(t)}$ and $l_{k+1}^{(t)}$ that bracket $l_{\rm GT}^{(t)}$ , with $\beta_k^{(t)} \leq \beta_{\rm GT}^{(t)} \leq \beta_{k+1}^{(t)}$.
    \STATE $u \gets \big(\beta^{(t)}_{\mathrm{GT}} - \beta^{(t)}_{k}\big) \big/
           \big(\beta^{(t)}_{{k+1}} - \beta^{(t)}_{k}\big)$ \hfill (Eq.~9)
    \STATE $\mathbf{a}^{\star(t)} \gets$ spherical interpolation of
           $\mathbf{a}^{(t)}_{k}$ and $\mathbf{a}^{(t)}_{{k+1}}$ by $u$
           \hfill (Eq.~10)
\ENDFOR
\STATE $\mathcal{L}_{\mathrm{level}} \gets \frac{1}{\sum_{t} p^{(t)}}
       \sum_{t \in \mathcal{T}} p^{(t)}
       \big|\hat{\beta}^{(t)} - \beta^{(t)}_{\mathrm{GT}}\big|$ \hfill (Eq.~17)
\STATE $\mathcal{L}_{\mathrm{align}} \gets \frac{1}{\sum_{t} p^{(t)}}
       \sum_{t \in \mathcal{T}} p^{(t)}
       \big(1 - \cos(\mathbf{z}^{(t)}, \mathbf{a}^{\star(t)})\big)$ \hfill (Eq.~12)
\STATE $\mathcal{L}_{\mathrm{total}} \gets
       \mathcal{L}_{\mathrm{CLIP}} + 0.1\,\mathcal{L}_{\mathrm{GT}}
     + \mathcal{L}_{\mathrm{conf}} + \mathcal{L}_{\mathrm{ord}}
     + 100\,\mathcal{L}_{\mathrm{align}} + \mathcal{L}_{\mathrm{level}}$ \hfill (Eq.~18)
\end{algorithmic}
\end{algorithm*}

\subsection{Construction and Properties of the Perceptual Severity Scale}
\label{sec:severity_scale}

Uniformly spaced physical degradation levels do not correspond to uniformly spaced perceptual changes. Following Eq.~(5), for each degradation type $t$ and physical level $l$, we compute $\beta_l^{(t)}$ as the mean LPIPS distance between clean images and their degraded counterparts over a fixed subset of DIV2K training images. The same clean images are used across all levels, and the resulting severity scores are precomputed before training and kept fixed. For downsampling, the degraded images are bicubically resized to the reference resolution before computing LPIPS. For a physical level not included in the tabulated grid, its severity is obtained by piecewise-linear interpolation between the two neighboring levels.

Table~\ref{tab:severity_range} reports the range of each type-specific severity scale and its Spearman rank correlation with degradation strength. All four types exhibit a strong ordinal relationship, with $\rho \geq 0.9987$, supporting the nearly monotonic behavior described in Sec.~3.3. Table~\ref{tab:severity_nonlinearity} further shows that uniformly spaced physical levels do not generally correspond to uniformly spaced perceptual severity. We therefore use the measured $\beta_l^{(t)}$, rather than the physical-level index, as the severity coordinate associated with each anchor.

\subsection{OD-CLIP Training Objective}
\label{sec.training_objective}

OD-CLIP keeps the CLIP image and text encoders frozen and introduces a trainable degradation encoder $E_d$, initialized from the CLIP visual backbone, whose hidden states are injected layer-wise into the frozen semantic encoder when processing the degraded image $x$. Consequently, although the semantic encoder remains frozen, $\mathcal{L}_{\mathrm{CLIP}}$ and $\mathcal{L}_{\mathrm{GT}}$ back-propagate through the injected features and update $E_d$. The clean image $y$ is processed by the frozen semantic encoder without degradation control.
Algorithm~\ref{alg:odclip_training} summarizes the training objective using the anchor levels in Sec.~A.1 and the perceptual severity scale in Sec.~A.2. When a ground-truth physical degradation level does not coincide with an anchor level, its target embedding is obtained by spherical interpolation between the two adjacent anchors that bracket it.
During training, the level and alignment losses are first averaged over samples in which the corresponding degradation type is present and then averaged across degradation types. $\mathcal{L}_{\mathrm{ord}}$ depends only on the anchors and is computed once per optimization step.

\begin{figure}[!t]
    \centering
    \includegraphics[width=1.0\linewidth]{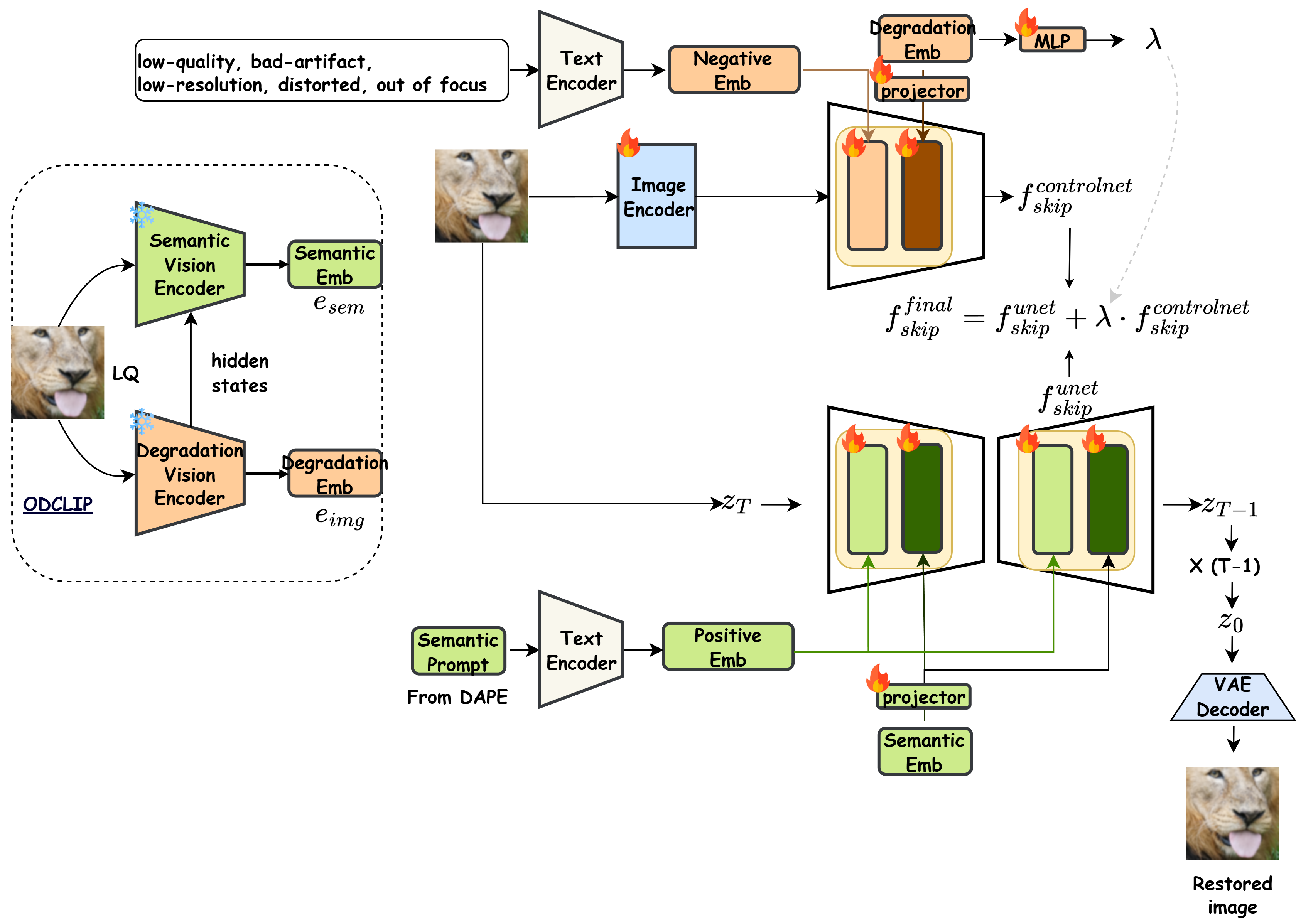}
    \caption{Integration of OD-CLIP into SeeSR \cite{seesr}. The frozen OD-CLIP encoder extracts a semantic embedding $\mathbf{e}_{\mathrm{sem}}$ and a degradation embedding $\mathbf{e}_{\mathrm{img}}$ from the low-quality input. The semantic embedding conditions the UNet, while the degradation embedding conditions the ControlNet and is further used to predict adaptive fusion weights $\boldsymbol{\lambda}$. Flames and snowflakes denote trainable and frozen modules, respectively.}
    \label{fig:seesr_odclip}

\end{figure}

\section{Integration into Diffusion-Based Blind SR}
In this section, we applied OD-CLIP to recent SOTA methods, SeeSR (SD 2.0)~\cite{seesr} and DiT4SR (SD 3.5)~\cite{dit4sr}.
Both are conditioned only on semantic content: their prompts are made robust to degradation, but the degradation itself is never represented, and restoration strength cannot adapt to it. 
We inject the OD-CLIP embeddings into both to supply this signal.
OD-CLIP is frozen throughout, and each backbone is fine-tuned with its original objective.

\subsection{Integration with SeeSR}
\label{sec:seesr_integration}

Fig.~\ref{fig:seesr_odclip} illustrates how OD-CLIP is integrated into
SeeSR \cite{seesr}. OD-CLIP remains frozen and extracts a semantic embedding
$\mathbf{e}_{\mathrm{sem}}$ and a degradation embedding
$\mathbf{e}_{\mathrm{img}}$ from the low-quality input. Two learnable
projection layers map these embeddings to the conditioning dimension
of the diffusion model. The semantic embedding conditions the UNet
image cross-attention layers, while the degradation embedding
conditions the corresponding ControlNet layers. In this way,
$\mathbf{e}_{\mathrm{sem}}$ provides content guidance, whereas
$\mathbf{e}_{\mathrm{img}}$ provides degradation-aware restoration
guidance.

The scalar severity estimates $\hat{\beta}^{(t)}$ are not explicitly passed to the generator. Instead, degradation type and severity information are conveyed through the ordinally structured degradation embedding $\mathbf{e}_{\mathrm{img}}$, which is directly shaped by the OD-CLIP training objectives.
The degradation embedding is additionally passed through a lightweight MLP to predict fusion weights $\boldsymbol{\lambda}$. These weights adaptively modulate the ControlNet features before they are added to the corresponding UNet features:
\begin{equation}
f_i^{\mathrm{final}}
=
f_i^{\mathrm{UNet}}
+
\lambda_i f_i^{\mathrm{ControlNet}}.
\label{eq:adaptive_fusion}
\end{equation}
This enables the strength of degradation guidance to adapt to the
input degradation.

\paragraph{Implementation details.} 
During training, the ControlNet, the two embedding projections, the UNet image cross-attention layers, and the fusion MLP are optimized using the standard diffusion $\epsilon$-prediction objective. OD-CLIP and the remaining diffusion components are kept frozen. We train the model on LSDIR~\cite{lsdir} at a resolution of $512\times512$ for $80$k iterations, using a learning rate of $5\times10^{-5}$, a batch size of $96$, and a text-dropout ratio of $0.5$ on a single NVIDIA RTX Pro 6000 GPU.
At inference, sampling starts from the noised low-quality latent and
uses $50$ denoising steps. For $4\times$ super-resolution, a
$128\times128$ input is processed at $512\times512$. We do not apply
classifier-free guidance. The ControlNet receives the fixed negative
quality prompt used during training, while the UNet receives the DAPE~\cite{seesr} semantic prompt.

\subsection{Integration with DiT4SR}
\label{sec:dit4sr_integration}

\begin{figure}[!t]
    \centering
    \includegraphics[width=1.0\linewidth]{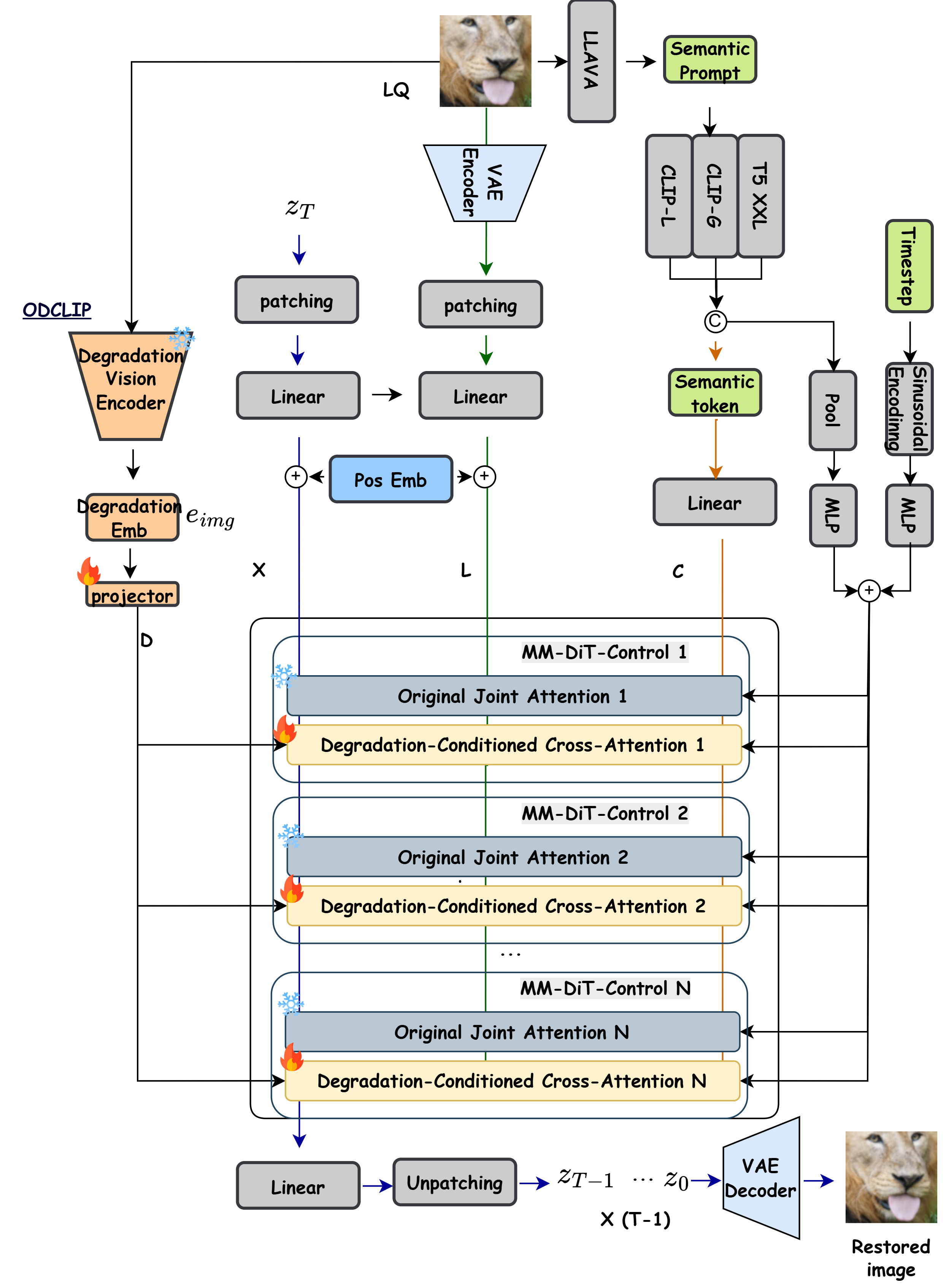}
    \caption{Integration of OD-CLIP into DiT4SR \cite{dit4sr}. The frozen OD-CLIP degradation encoder extracts a degradation embedding $\mathbf{e}_{\mathrm{img}}$ from the low-quality input, which is projected into degradation tokens $\mathbf{D}$. In each MMDiT block, these tokens are introduced through a degradation-conditioned cross-attention layer placed after the original joint-attention update. The pretrained DiT4SR backbone remains frozen, while only the degradation-token projector and the newly added cross-attention modules are optimized. Snowflakes and flames denote frozen and trainable modules, respectively.}
    \label{fig:dit4sr_odclip}

\end{figure}

Unlike the SeeSR \cite{seesr} integration, the DiT4SR \cite{dit4sr} variant is adapted from a pretrained DiT4SR checkpoint by introducing a lightweight degradation-conditioning pathway. Both the pretrained DiT4SR backbone and OD-CLIP remain frozen, and only the newly added conditioning modules are optimized.

Given a low-quality input $x$, the frozen OD-CLIP encoder extracts the degradation embedding $\mathbf{e}_{\mathrm{img}}$. We use only the degradation branch of OD-CLIP, without introducing its semantic embedding. A learnable projection converts $\mathbf{e}_{\mathrm{img}}$ into degradation tokens: \begin{equation} \mathbf{D} = \phi\left(\mathbf{e}_{\mathrm{img}}\right), \label{eq:dit4sr_deg_tokens} \end{equation} which provide the degradation condition for the MMDiT blocks.

We introduce a degradation-conditioned cross-attention layer after the original joint-attention update in each MMDiT block. Let $\mathbf{h}_i$ denote the HR image tokens after the original joint attention and its residual connection in the $i$-th block. These updated HR tokens serve as queries, while the degradation tokens provide the keys and values: \begin{equation}
\begin{aligned}
\widetilde{\mathbf{h}}_i
&=
\mathbf{h}_i
+
\tanh(\gamma_i)\,
\operatorname{CrossAttn}_i
\left(
\begin{aligned}
Q &= \mathbf{h}_i,\\
K &= \mathbf{D},\quad
V = \mathbf{D}
\end{aligned}
\right).
\end{aligned}
\label{eq:dit4sr_deg_attn}
\end{equation} where $\gamma_i$ is a learnable scalar gate. The resulting degradation-conditioned tokens $\widetilde{\mathbf{h}}_i$ are then passed to the original MLP sublayer.

The added cross-attention uses projections independent of the original joint-attention projections and modifies only the HR image-token stream. Each gate is initialized to zero, so the additional pathway has no effect at the beginning of training and the model initially preserves the behavior of the pretrained DiT4SR checkpoint.

\paragraph{Implementation details.}
During training, the DiT4SR backbone and OD-CLIP are kept frozen. Only the degradation-token projection, the added cross-attention layers, and their gates are optimized using the original flow-matching objective of DiT4SR.
We train the added modules for $10$k steps using high-resolution images from DIV2K and Flickr2K, with low-quality inputs synthesized using the Real-ESRGAN degradation pipeline. Training uses AdamW, an batch size of $64$ on a single NVIDIA RTX Pro 6000 GPU, and a learning rate of $2.5\times10^{-4}$ for the newly added modules. The text-prompt dropout ratio is set to $0.2$. At inference, we use $40$ diffusion steps and a classifier-free guidance scale of $8.0$. Sampling starts from the noised low-quality latent, and AdaIN is applied for color alignment.

\section{Experiment Settings and Implementation Details}

\begin{algorithm}[!t]
\caption{Mixed-Degradation Synthesis Pipeline}
\label{alg:multi_deg_syn}
\textbf{Input:} Clean image $y$ \\
\textbf{Output:} Degraded image $x$, degradation prompt
$c_{\mathrm{deg}}$

$x \leftarrow y$ \\
$c_{\mathrm{deg}} \leftarrow \emptyset$

\textbf{while} $c_{\mathrm{deg}} = \emptyset$ \textbf{do}

\quad \textbf{for}
$t \in \{\text{blur},\text{downsampling},\text{noise},\text{JPEG}\}$
\textbf{do}

\quad\quad sample $p \sim \mathcal{U}(0,1)$

\quad\quad \textbf{if} $p < 0.5$ \textbf{then}

\quad\quad\quad sample degradation parameter $\theta_t$

\quad\quad\quad $x \leftarrow D_t(x,\theta_t)$

\quad\quad\quad
$c_{\mathrm{deg}}
\leftarrow
\operatorname{Concat}
(c_{\mathrm{deg}},t(\theta_t))$

\quad\quad \textbf{end if}

\quad \textbf{end for}

\textbf{end while}

\textbf{return} $x,c_{\mathrm{deg}}$
\end{algorithm}

\subsection{Degradation Synthesis}
\label{sec:degradation_synthesis}

Following the degradation settings described in the main paper, we synthesize low-quality images from clean inputs using four degradation types: Gaussian blur, downsampling, Gaussian noise, and JPEG compression. Their physical parameter ranges and discretization intervals are reported in Sec.~\ref{sec.pdpg_and_anchor_const}.
For single-degradation samples, exactly one degradation operator is applied to each clean image. Multiple degraded versions are generated at different physical levels, allowing the model to learn fine-grained severity variations while keeping the image content fixed. For mixed-degradation samples, each degradation type is independently selected with probability $0.5$, while ensuring that at least one type is selected. The parameter of each selected degradation is sampled from its corresponding range, and the selected operators are applied sequentially in the order of Gaussian blur, downsampling, Gaussian noise, and JPEG compression. Along with the degraded image, we store a textual degradation record $c_{\mathrm{deg}}$ containing the applied degradation types and their sampled parameters. The complete synthesis procedure is summarized in Algorithm~\ref{alg:multi_deg_syn}.

\subsection{Choice of the Perceptual Severity Metric}
\label{sec:severity_metric_choice}

\begin{figure}[!t]
    \centering
    \includegraphics[width=1.0\linewidth]{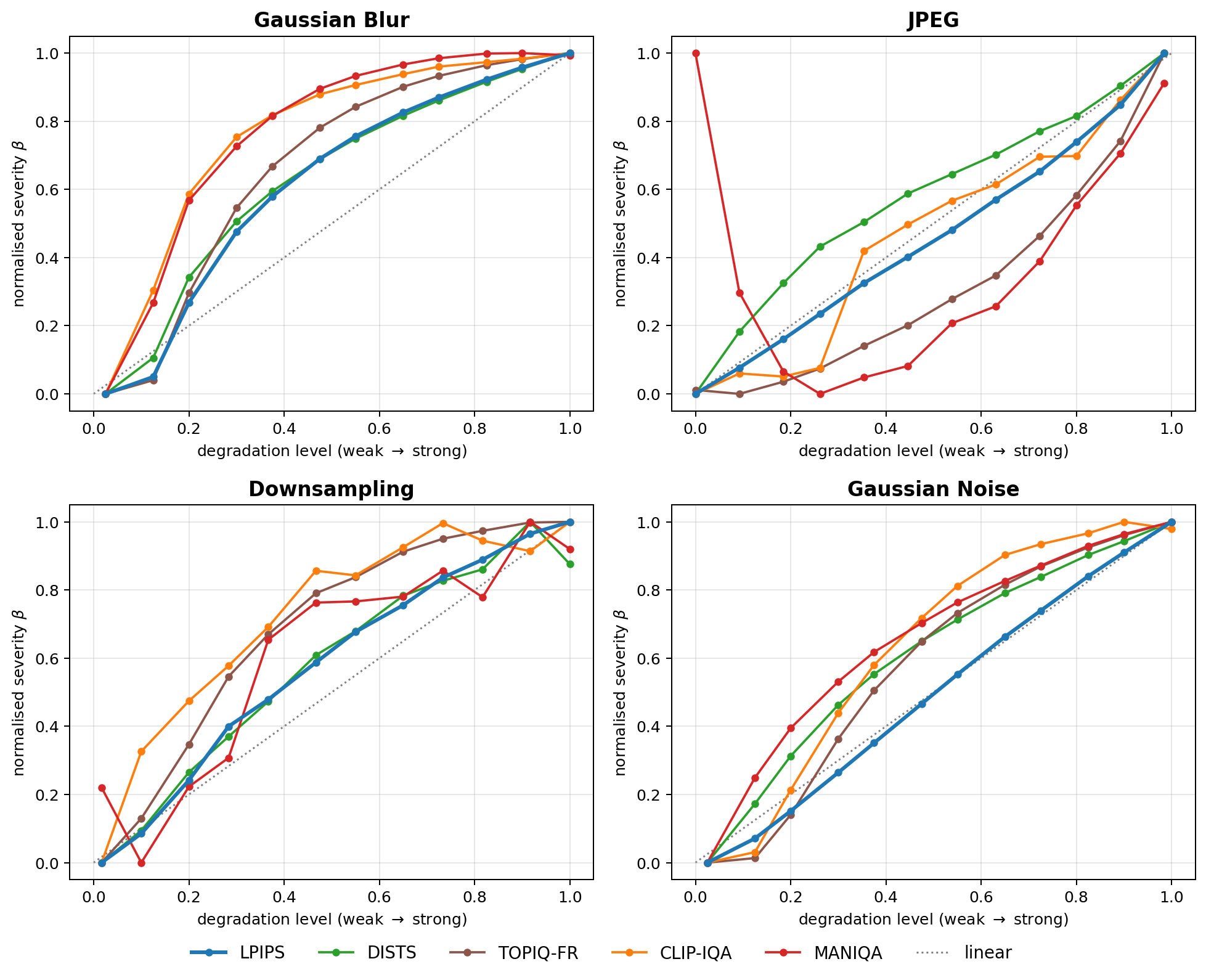}
    \caption{
    Comparison of candidate metrics for defining perceptual degradation severity. The responses are oriented such that larger values indicate stronger degradation and are normalized independently for each degradation type. Among the evaluated metrics, LPIPS exhibits the most consistent smooth and monotonic progression across Gaussian blur, JPEG compression, downsampling, and Gaussian noise, motivating its use to define $\beta_l^{(t)}$. The dotted line indicates a linear response to the normalized physical degradation level.
    }
    \label{fig:metric_choice}
\end{figure}

Equation~(5) defines the perceptual severity coordinate using LPIPS. To justify this choice, we compare LPIPS with four alternative image-quality metrics: DISTS \cite{dists}, TOPIQ-FR \cite{topiq}, CLIP-IQA \cite{clipiqa}, and MANIQA \cite{maniqa}. For each degradation type, the metric responses are oriented such that larger values indicate stronger degradation and are normalized to $[0,1]$. The comparison therefore focuses on the shape and ordering of the responses rather than their original numerical scales. A suitable severity metric should provide a smooth and monotonic response as the physical degradation becomes stronger, since local reversals or early saturation would assign similar or inconsistent severity values to distinct degradation levels. As shown in Fig.~\ref{fig:metric_choice}, LPIPS provides the most consistent response across all four degradation types. Its value increases smoothly and monotonically from weak to strong degradation for Gaussian blur, JPEG compression, downsampling, and Gaussian noise. In comparison, DISTS exhibits a local reversal at the strongest downsampling levels. TOPIQ-FR and MANIQA show pronounced non-monotonic responses for JPEG compression and additional irregularities under downsampling. CLIP-IQA increases rapidly and saturates early for Gaussian blur, while also showing less uniform responses for JPEG compression and downsampling. We therefore adopt LPIPS to construct the perceptual severity coordinate $\beta_l^{(t)}$ in Eq.~(5).

\section{Additional Quantitative Experiments}

\begin{figure}[!t]
    \centering
    \includegraphics[width=1.0\linewidth]{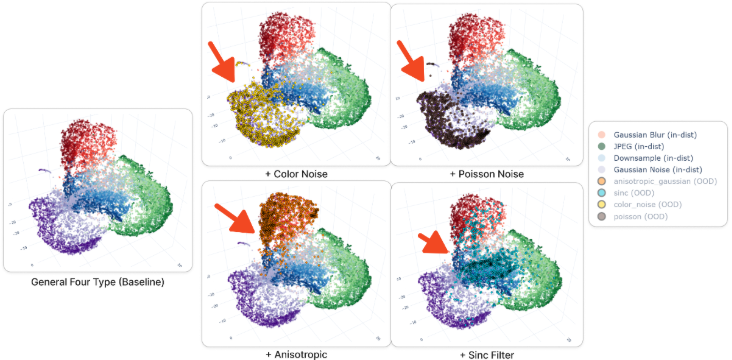}
    \caption{
    3D t-SNE visualization of degradation embeddings. Although anisotropic Gaussian blur, 2D sinc filtering, Poisson noise, and color noise are not included during training, their embeddings are still mapped close to related degradation clusters in the learned degradation space.}
    \label{fig:tsne_ood}
\end{figure}

\subsection{Generalization to Unseen Degradations}
\label{sec:ood_degradation}

While OD-CLIP is trained using only four degradation types, i.e., Gaussian blur, Gaussian noise, downsampling, and JPEG compression, we further examine whether it can generalize to unseen degradations beyond the training set. To this end, we analyze both the degradation embedding distribution and the predicted degradation parameters on several out-of-distribution degradations, including anisotropic Gaussian blur, 2D sinc filtering, Poisson noise, and color noise.
Fig.~\ref{fig:tsne_ood} shows the 3D t-SNE visualization of the degradation embeddings. Although these degradations are not included during training, their embeddings are still mapped to related regions in the learned degradation space. This indicates that OD-CLIP, instead of just memorizing training operators, learns a structured degradation representation that can generalize to unseen but semantically similar degradations.

\begin{figure}[!t]
    \centering
    \includegraphics[width=1.0\linewidth]{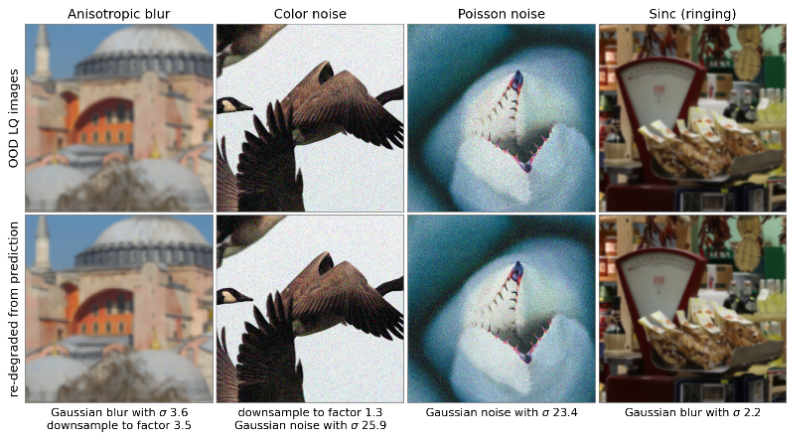}
    \caption{
    Qualitative re-degradation results on unseen degradation types. The top row shows OOD low-quality inputs, while the bottom row shows the corresponding clean images re-degraded using the in-vocabulary types and physical parameters predicted by OD-CLIP. The predicted configurations reproduce the main visual characteristics of the unseen degradations.
    }
    \label{fig:visualize_ood}
\end{figure}

\paragraph{Image-space re-degradation.}
To complement the embedding space analysis, we examine whether OD-CLIP can approximate unseen degradations using the degradation axes learned from the four training types. For each degraded input, OD-CLIP predicts the active degradation types and their corresponding severities. We map the predicted severities back to physical parameters and apply the resulting degradation configuration to the paired clean image using the same operator order as in the training synthesis pipeline.
As shown in Fig.~\ref{fig:visualize_ood}, the re-degraded images reproduce the main visual characteristics and overall strength of the corresponding unseen degradations. OD-CLIP maps each unseen degradation to its closest single or mixed combination of known degradation axes, producing visually similar observations even though these degradation types were not included during training. These results suggest that the learned degradation representation can meaningfully approximate unseen but perceptually related degradations beyond the predefined training types.

\subsection{Qualitative Re-degradation Analysis}
\label{sec:redegradation}

\begin{figure}[!t]
    \centering
    \includegraphics[width=1.0\linewidth]{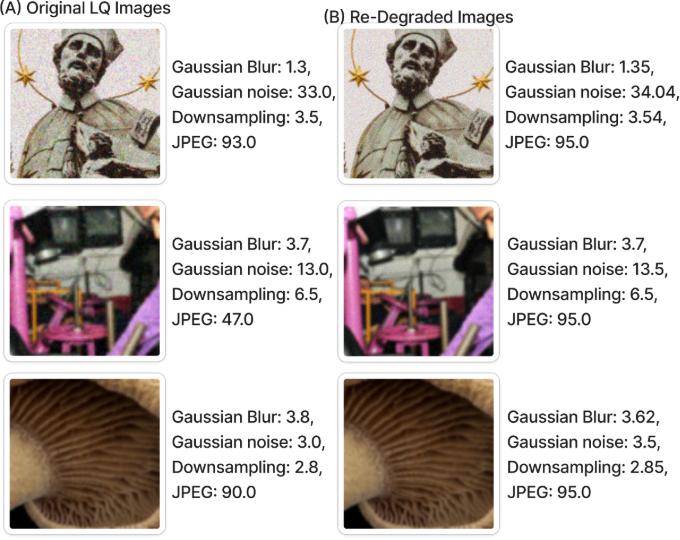}
    \caption{
    Qualitative re-degradation results. OD-CLIP first estimates the active degradation types and their physical levels from each original low-quality image. The predicted degradations are then applied to the corresponding ground-truth image using the same operator order as the dataset synthesis pipeline. The re-degraded images reproduce visual characteristics similar to the original observations. Ground-truth and estimated degradation parameters are shown beside each image.}
    \label{fig:redegradation}
\end{figure}

We further validate the degradation estimates of OD-CLIP through re-degradation. Given a paired ground-truth image $y$ and its low-quality image $x$, OD-CLIP predicts the active degradation types and their physical levels from $x$. We then apply the predicted degradations to $y$ in the same order used for dataset synthesis, namely Gaussian blur, downsampling, Gaussian noise, and JPEG compression, while omitting types predicted as absent. Fig.~\ref{fig:redegradation} compares the original and re-degraded low-quality images. The re-degraded results reproduce similar blur, noise, resolution loss, and compression characteristics, and the estimated parameters remain close to the ground-truth synthesis settings. This provides an intuitive end-to-end validation of the mixed-degradation representations learned by OD-CLIP.

\subsection{Comparison of Degradation-Aware Conditioning}
\label{sec:comparison_daclip_sr}

To isolate the effect of degradation representation, we compare DA-CLIP and OD-CLIP under the same SeeSR architecture and training protocol. Specifically, we retain the training data, optimization settings, conditioning interfaces, and inference configuration, and only replace the degradation-aware encoder. Since DA-CLIP also provides degradation-conditioned features for image restoration, this controlled comparison directly evaluates whether the ordinal degradation representation learned by OD-CLIP provides more effective conditioning for blind SR.

As shown in Table~\ref{tab:daclip_comparison}, OD-CLIP achieves the highest PSNR and SSIM on all three benchmarks. Compared with DA-CLIP, it improves PSNR by $0.57$, $1.19$, and $0.60$ dB on DIV2K-Val, DRealSR, and RealSR, respectively, with corresponding SSIM gains of $0.0159$, $0.0318$, and $0.0146$. These consistent improvements indicate that OD-CLIP provides more reliable degradation information for reconstructing image structures and content.

The difference is more evident on real-world degradations. On both DRealSR and RealSR, OD-CLIP also achieves lower LPIPS and DISTS than DA-CLIP, showing that its fidelity improvements are accompanied by better perceptual similarity. In contrast, the effect of DA-CLIP is less consistent across datasets: it improves several metrics on DIV2K-Val, but provides limited or negative gains on DRealSR. Overall, the results demonstrate that explicitly organizing degradation representations according to perceptual severity leads to more stable and effective conditioning than using degradation-aware features without ordinal supervision.

\begin{table}[t]
    \centering
    \small
    \setlength{\tabcolsep}{1mm}
    \begin{tabular}{l ccccc}
    \toprule
    & \multicolumn{2}{c}{\emph{Fidelity}} & \multicolumn{2}{c}{\emph{Perceptual}} &  \emph{Distribution}\\
    \cmidrule(lr){2-3}  \cmidrule(lr){4-5} \cmidrule(lr){6-6}
    \textbf{Method} & \textbf{PSNR}$\uparrow$ & \textbf{SSIM}$\uparrow$ & \textbf{LPIPS}$\downarrow$ & \textbf{DISTS}$\downarrow$ & \textbf{FID}$\downarrow$ \\
    \midrule
    \multicolumn{6}{c}{\textbf{DIV2K-Val} \cite{div2k}} \\
    \midrule
    
    SeeSR & 23.7250 & 0.6056 & 0.3194 & 0.1966 & 25.8199\\
    + DACLIP & \second{23.8668} & \second{0.6162} & \best{0.3034} & \best{0.1893} & \best{24.6240}  \\
    + ODCLIP & \best{24.4334} & \best{0.6321} & \second{0.3096} & \second{0.1912} & \second{24.8198}\\
    \midrule
    \multicolumn{6}{c}{\textbf{DRealSR} \cite{drealsr}} \\
    \midrule
    SeeSR & \second{28.1430} & \second{0.7711} & 0.3142 & \second{0.2299} & \best{146.9934} \\
    + DA-CLIP & 27.6833 & 0.7691 & \second{0.3134} & 0.2323 & 151.0608 \\
    + OD-CLIP & \best{28.8697} & \best{0.8009} & \best{0.2963} & \best{0.2227} & \second{148.4187} \\
    \midrule
    \multicolumn{6}{c}{\textbf{RealSR} \cite{realsr}} \\
    \midrule
   
    SeeSR & 25.2082 & \second{0.7215} & 0.3004 & 0.2218 & \best{124.9782} \\
    + DA-CLIP & \second{25.2789} & 0.7324 & \second{0.2903} & \second{0.2209} & \second{131.6345} \\
    + OD-CLIP & \best{25.8823} & \best{0.7470} & \best{0.2860} & \best{0.2206} & 135.3407 \\
    \bottomrule
    \end{tabular}
    \caption{Quantitative evaluation on DIV2K-Val, DRealSR, and RealSR benchmarks. The best and second-best results are highlighted in \textcolor{red}{Red} and \textcolor{blue}{Blue}, respectively.
    }
    \label{tab:daclip_comparison}
\end{table}

\begin{table}[t]
\centering
\small
\setlength{\tabcolsep}{1mm}
\setlength{\tabcolsep}{6pt}
\renewcommand{\arraystretch}{1.1}
\begin{tabular}{cccccc}
\toprule
$\Delta l$ & top-$k$ & Type Acc$\uparrow$ & MAE$\downarrow$ & SROCC$\uparrow$ & PCC$\uparrow$ \\
\midrule
\textbf{1} & \textbf{2} & \textbf{\textcolor{red}{0.8483}} & \textit{\textcolor{blue}{0.1218}} & \textbf{\textcolor{red}{0.8875}} & \textbf{\textcolor{red}{0.8668}} \\
\midrule
1  & 1   & 0.7843 & 0.2254 & 0.5019 & 0.5067 \\
1  & 5   & 0.8045 & 0.1249 & 0.7633 & 0.7254 \\
1  & ALL & 0.7997 & \textbf{\textcolor{red}{0.1188}} & 0.7935 & 0.7344 \\
\midrule
5  & 2 & \textit{\textcolor{blue}{0.8255}} & 0.1390 & 0.7560 & 0.6945 \\
10 & 2 & 0.7212 & 0.1406 & 0.8195 & 0.7551 \\
20 & 2 & 0.7828 & 0.1459 & \textit{\textcolor{blue}{0.8342}} & \textit{\textcolor{blue}{0.7794}} \\
\bottomrule
\end{tabular}
\caption{Ablation of the index stride $\Delta l$ used to select anchors from the enumerated physical levels and the number $K$ of highest-similarity anchors used for severity estimation. We vary one factor at a time while keeping the other at its default value ($\Delta l=1$ and $K=2$). The best and second-best results are highlighted in \textcolor{red}{Red} and \textcolor{blue}{Blue}, respectively.}
\label{tab:ablation_anchor}
\end{table}

\section{Ablation Studies and Analysis}

\subsection{Analysis of Anchor Configuration}
\label{sec:ablation_anchor}

Table~\ref{tab:ablation_anchor} examines two components of the anchor configuration: the index stride $\Delta l$ used to select anchors from the enumerated physical levels, and the number $K$ of highest-similarity anchors used for severity estimation. We vary one factor at a time while keeping the other at its default value ($\Delta l=1$ and $K=2$).

With $\Delta l=1$, using the two highest-similarity anchors achieves the highest Type Accuracy, SROCC, and PCC. Using only one anchor causes a substantial performance drop, indicating that a single-anchor estimate is insufficient for capturing continuous severity. Increasing $K$ to $5$ or using all anchors improves over $K=1$, but remains inferior to $K=2$ in Type Accuracy and correlation metrics. Although using all anchors gives a slightly lower MAE ($0.1188$ versus $0.1218$), its SROCC and PCC are considerably lower. Overall, $K=2$ provides the most balanced performance.
We further vary the index stride $\Delta l$ while fixing $K=2$. The default setting $\Delta l=1$, which selects every enumerated physical level, achieves the best performance across all four metrics. Coarser anchor selection generally reduces both type classification and severity-estimation performance, showing the benefit of retaining fine-grained anchor levels.

\begin{figure}[t]
\centering
\includegraphics[width=\linewidth]{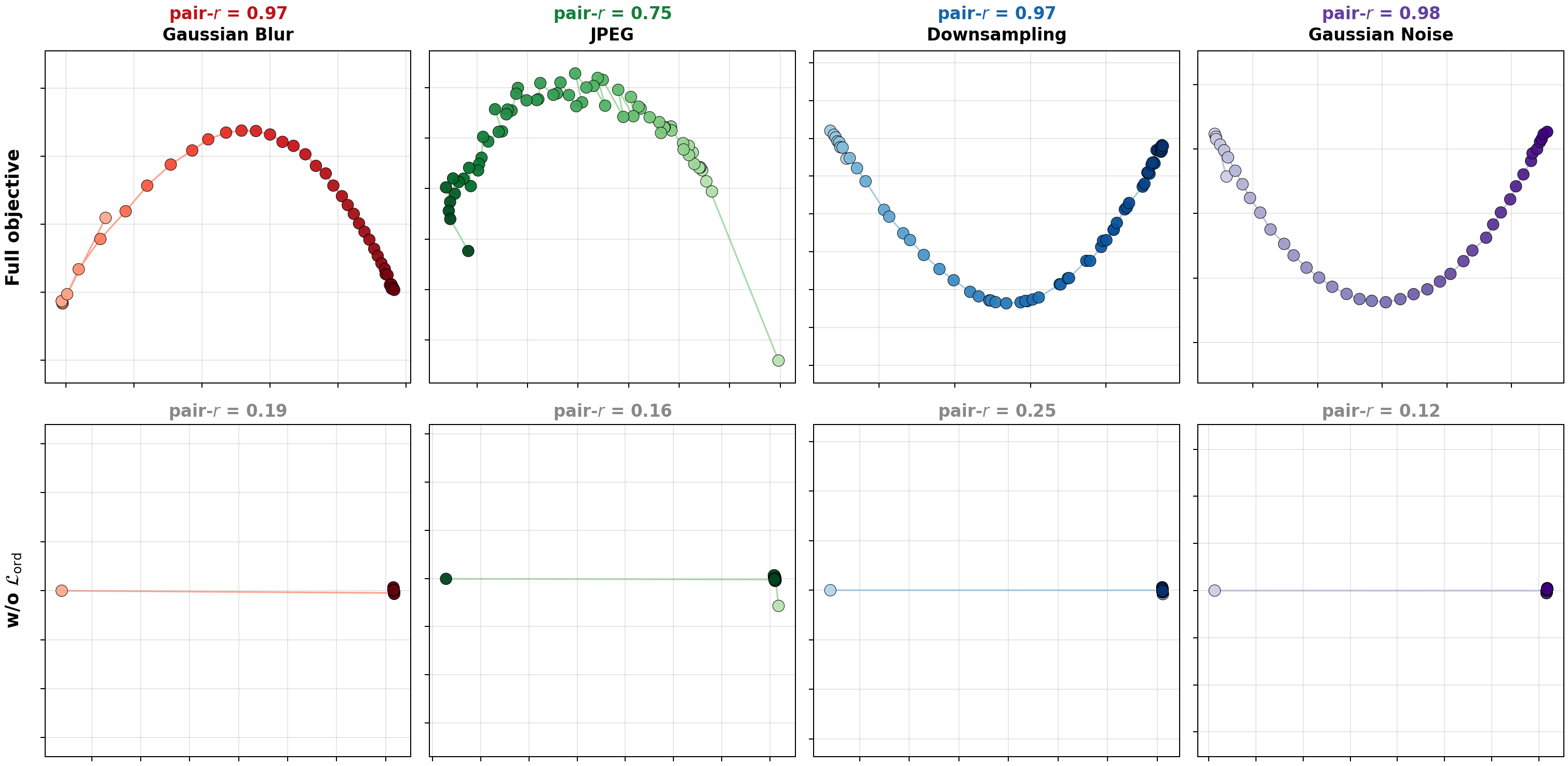}
\caption{Effect of the ordinal metric loss on the learned anchor geometry. Each column visualizes the severity anchors of one degradation type using PCA. Anchors are colored by their perceptual severity $\beta_l^{(t)}$ and connected in severity order. The top and bottom rows show the full objective and the model trained without $\mathcal{L}_{\mathrm{ord}}$, respectively. With $\mathcal{L}_{\mathrm{ord}}$, the anchors form ordered trajectories whose relative spacing follows the perceptual severity scale more closely. Removing the loss produces more compact configurations with a weaker correspondence between anchor geometry and severity.}
\label{fig:ord-ablation}
\end{figure}

\subsection{Effect of the Ordinal Metric Loss on Anchor Geometry}
\label{sec:ordinal_anchor_geometry}

Each severity anchor is constructed as the sum of a type embedding and a learnable level-specific shift, as defined in Eq.~(6). Although multiple objectives provide supervision to the anchors, $\mathcal{L}_{\mathrm{ord}}$ is the component that explicitly regularizes their within-type pairwise geometry. In particular, it encourages the normalized distance between two anchors, $D_{ij}^{(t)}$, to follow their corresponding perceptual severity difference, $B_{ij}^{(t)}$. Fig.~\ref{fig:ord-ablation} compares the learned anchor geometry under the full objective and after removing $\mathcal{L}_{\mathrm{ord}}$. With the ordinal metric loss, the anchors form smooth trajectories ordered by perceptual severity, and their relative spacing more closely follows the severity scale. Without $\mathcal{L}_{\mathrm{ord}}$, the anchors become substantially more compact and exhibit a weaker correspondence between embedding distance and perceptual severity difference. These observations show that $\mathcal{L}_{\mathrm{ord}}$ provides explicit geometric supervision for organizing the within-type anchor space according to perceptual severity, complementing the sample-level alignment and severity-estimation objectives.

\section{Additional Qualitative Results}

\begin{figure*}[t]
    \centering
    \includegraphics[width=1.0\linewidth]{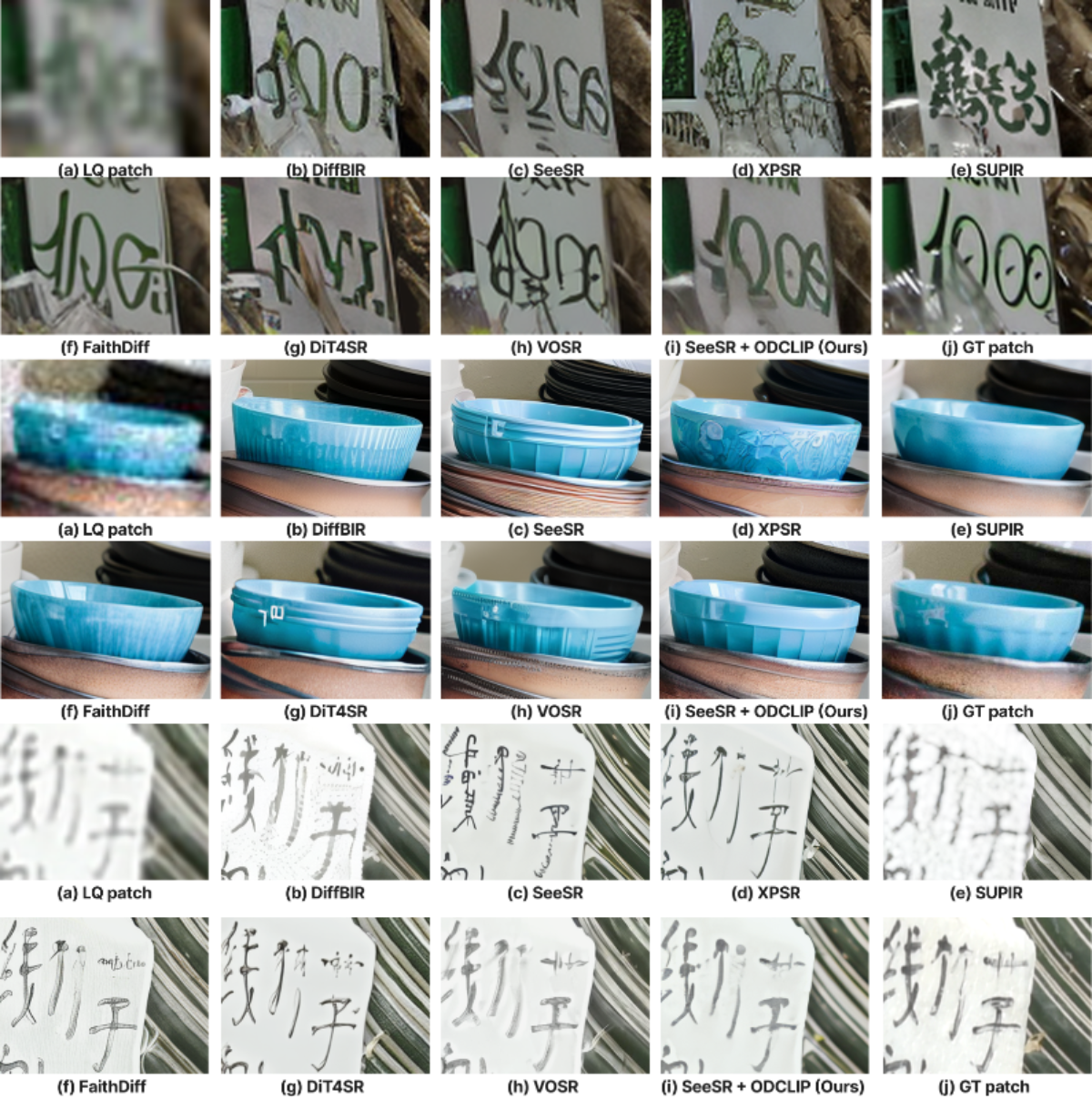}
    \caption{
    Qualitative comparisons with SOTA methods. With OD-CLIP, our method achieves better fidelity and content structure. \textbf{(Zoom in for a better view.)}
    }
    \label{fig:stage2_supp_visualize1}
\end{figure*}

\begin{figure*}[t]
    \centering
    \includegraphics[width=1.0\linewidth]{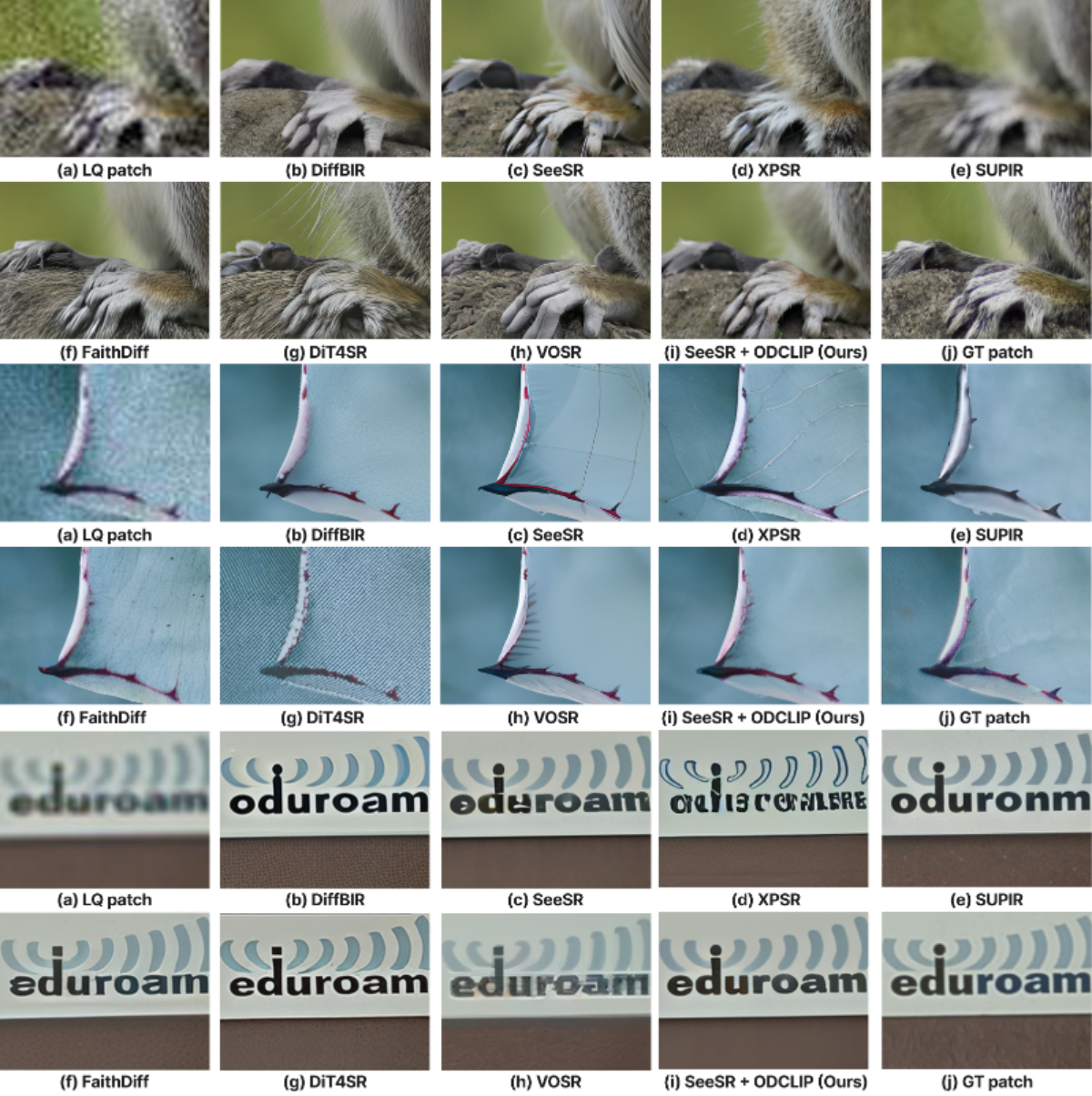}
    \caption{
    Qualitative comparisons with SOTA methods. With OD-CLIP, our method achieves better fidelity and content structure. \textbf{(Zoom in for a better view.)}
    }
    \label{fig:stage2_supp_visualize2}
\end{figure*}

Additional visualization results of our method (OD-CLIP) on test datasets are shown in Figs~\ref{fig:stage2_supp_visualize1} and \ref{fig:stage2_supp_visualize2}.

\end{document}